\documentclass{article}

\usepackage{microtype}
\usepackage{graphicx}
\usepackage{subfigure}
\usepackage{booktabs} %

\usepackage{hyperref}
\usepackage{url}
\usepackage{amsfonts}       %
\usepackage{nicefrac}       %
\usepackage{xcolor}         %
\usepackage{multirow}
\usepackage{enumitem}
\usepackage{adjustbox}
\usepackage{colortbl}
\usepackage{nicematrix}
\definecolor{mygray}{gray}{0.85}
\definecolor{softgray}{rgb}{0.9, 0.9, 0.9}
\definecolor{softblue}{rgb}{0.88, 0.92, 1.0}
\definecolor{softgreen}{rgb}{0.88, 1.0, 0.88}
\definecolor{softyellow}{rgb}{1.0, 1.0, 0.88}
\definecolor{softred}{rgb}{1.0, 0.88, 0.88}
\definecolor{softpink}{rgb}{1.0, 0.88, 0.94}

\usepackage[accepted]{icml2024}

\usepackage{amsmath}
\usepackage{amssymb}
\usepackage{mathtools}
\usepackage{amsthm}

\usepackage[capitalize,noabbrev]{cleveref}

\theoremstyle{plain}

\theoremstyle{definition}

\theoremstyle{remark}

\usepackage[textsize=tiny]{todonotes}

\usepackage{color}
\usepackage{ifthen}

\definecolor{zhiqiu_color}{rgb}{0,0.5,1}
\definecolor{deva_color}{rgb}{0.2,.64,0}
\definecolor{deepak_color}{rgb}{1,0,1}
\definecolor{xinyue_color}{rgb}{1,0,1}
\definecolor{pengchuan_color}{rgb}{1,0,1}

\newif\ifsubmit
\submitfalse
\ifsubmit
    \newcommand{\zhiqiu}[1]{}
    \newcommand{\deva}[1]{}
    \newcommand{\deepak}[1]{}
    \newcommand{\xinyue}[1]{}
    \newcommand{\pengchuan}[1]{}
\else
    \newcommand{\zhiqiu}[1]{\textsf{\textcolor{zhiqiu_color}{[{\bf Zhiqiu}: #1]}}}
    
    \newcommand{\deepak}[1]{\textsf{\textcolor{deepak_color}{[{\bf Deepak}: #1]}}}
    \newcommand{\xinyue}[1]{\textsf{\textcolor{xinyue_color}{[{\bf Xinyue}: #1]}}}
    \newcommand{\pengchuan}[1]{\textsf{\textcolor{pengchuan_color}{[{\bf Pengchuan}: #1]}}}
\fi

\usepackage{amsmath}
\usepackage{bbm}

\icmltitlerunning{Revisiting the Role of Language Priors in Vision-Language Models}

\begin{document}

\twocolumn[
\icmltitle{Revisiting the Role of Language Priors in Vision-Language Models}

\icmlsetsymbol{equal}{*}

\begin{icmlauthorlist}
\icmlauthor{Zhiqiu Lin}{equal,cmu}
\icmlauthor{Xinyue Chen}{equal,cmu}
\icmlauthor{Deepak Pathak}{cmu}
\icmlauthor{Pengchuan Zhang}{meta}
\icmlauthor{Deva Ramanan}{cmu}
\end{icmlauthorlist}

\icmlaffiliation{cmu}{CMU}
\icmlaffiliation{meta}{Meta}

\icmlcorrespondingauthor{Zhiqiu Lin}{zhiqiul@andrew.cmu.edu}

\icmlkeywords{Machine Learning, Computer Vision, Natural Language Processing, Vision-Language Models, Cross-Modal Retrieval, ICML}

\vskip 0.3in
]

\printAffiliationsAndNotice{\icmlEqualContribution} %

\begin{abstract}
Vision-language models (VLMs) are impactful in part because they can be applied to a variety of visual understanding tasks in a zero-shot fashion, without any fine-tuning. We study \textit{generative VLMs} that are trained for next-word generation given an image. We explore their zero-shot performance on the illustrative task of image-text retrieval across nine popular vision-language benchmarks. Our first observation is that they can be repurposed for discriminative tasks (such as image-text retrieval) by simply computing the match score of generating a particular text string given an image. We call this probabilistic score the {\em Visual Generative Pre-Training Score} (VisualGPTScore). While the VisualGPTScore produces near-perfect accuracy on some retrieval benchmarks, it yields poor accuracy on others. We analyze this behavior through a probabilistic lens, pointing out that some benchmarks inadvertently capture unnatural language distributions by creating adversarial but unlikely text captions. In fact, we demonstrate that even a ``blind'' language model that ignores any image evidence can sometimes outperform all prior art, reminiscent of similar challenges faced by the visual-question answering (VQA) community many years ago. We derive a probabilistic post-processing scheme that controls for the amount of linguistic bias in generative VLMs at test time without having to retrain or fine-tune the model. We show that the VisualGPTScore, when appropriately debiased, is a strong zero-shot baseline for vision-language understanding, oftentimes producing state-of-the-art accuracy.
\end{abstract}

\begin{figure*}[h]
\centering
    \scalebox{0.77}{
    \begin{tabular}{l l | r r}
        \multicolumn{2}{c}{{\Large Scenario 1 
        }} & \multicolumn{2}{c}{{\Large Scenario 2 
        }}\\
        \includegraphics[width=0.40\textwidth]{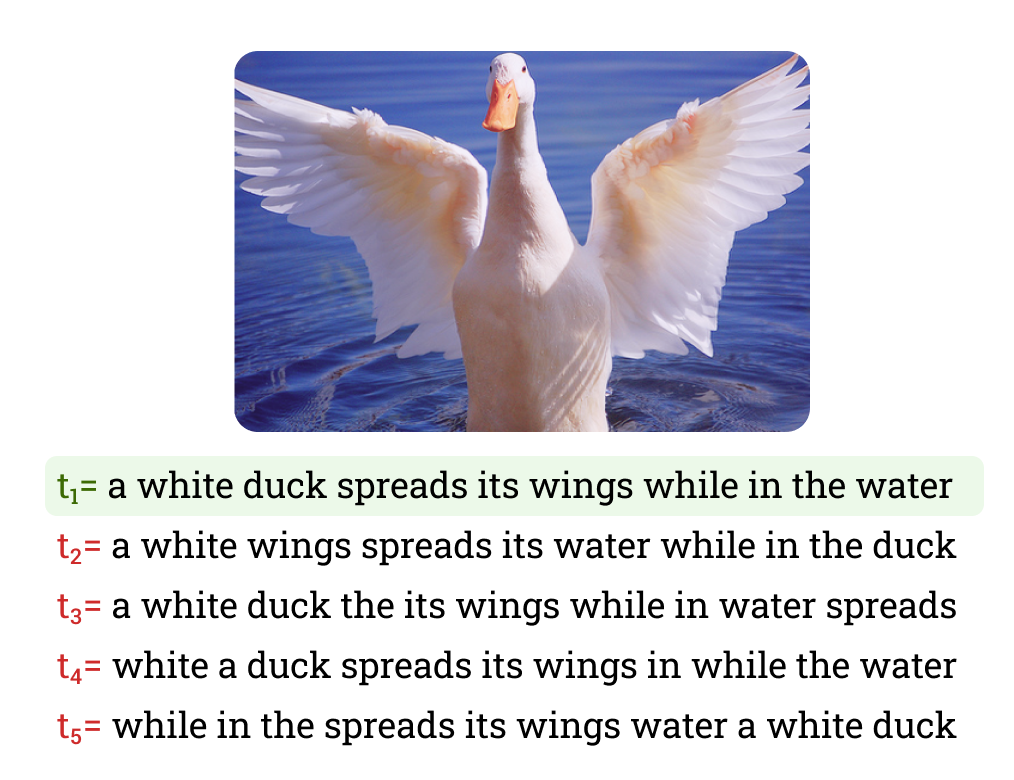} &
        \includegraphics[width=0.21\textwidth]{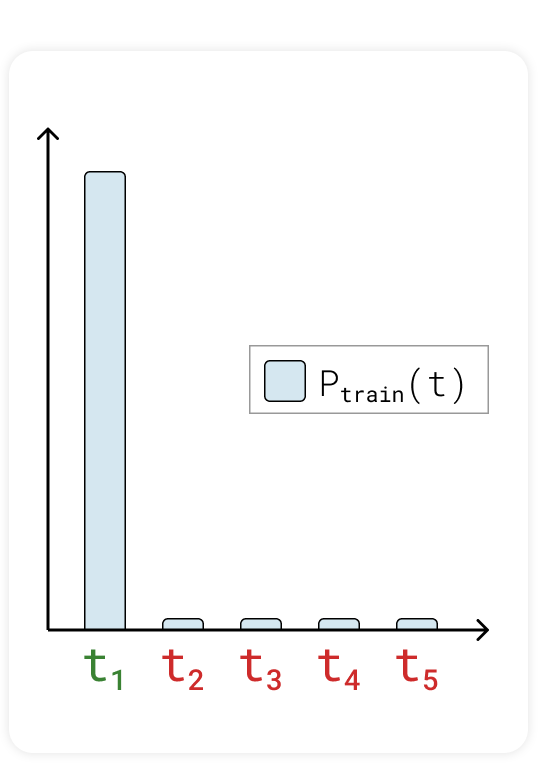} &
        \includegraphics[width=0.275\textwidth]{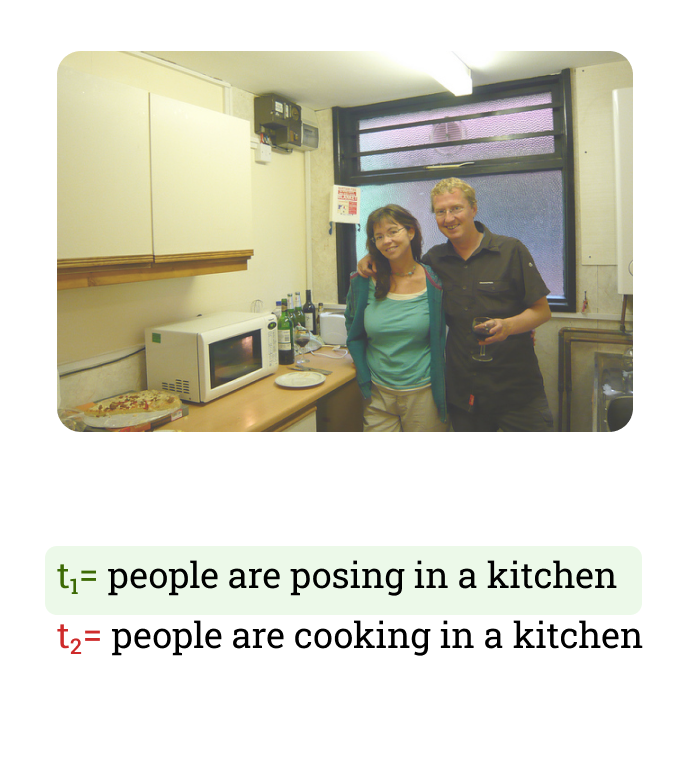} &
        \includegraphics[width=0.21\textwidth]{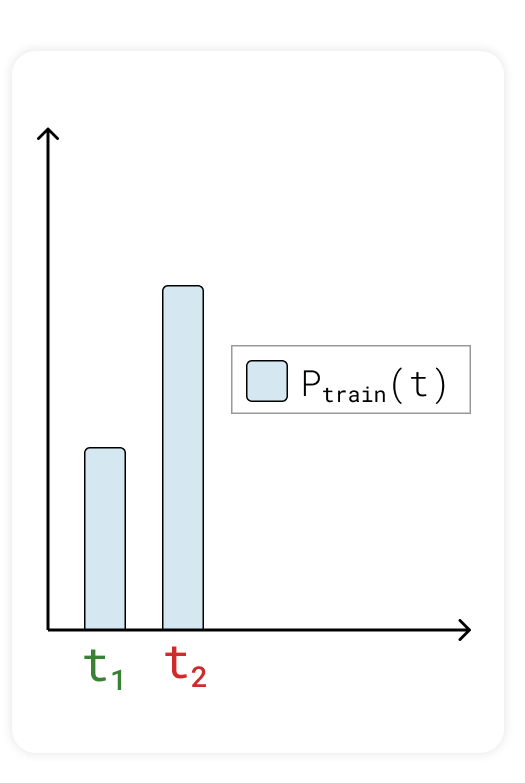} \\
    \end{tabular}
    }
    \caption{\small {\bf Two train-test shifts encountered in image-to-text retrieval tasks.} Scenario 1 ({\bf left}) constructs negative captions by shuffling words in the true caption (as in ARO-Flickr \cite{aro}), but this produces implausible text such as ``{\tt white a duck spreads its wings in while the water}". Here, exploiting the language bias of the training set will help since it will downweight the match score for such implausible negative captions. In fact, we show that a blind language-only model can easily identify the correct caption.  Scenario 2 ({\bf right}) constructs negative captions that are curated to be plausible (as in SugarCrepe \cite{sugarcrepe}). Here, the language bias of the training set may hurt, since it will prefer to match common captions that score well under the language prior; i.e., the incorrect caption of ``{\tt people are cooking in a kitchen}" is slightly more likely than the true caption of ``{\tt people are posing in a kitchen}" under the language prior, and so removing the language bias improves performance. We present simple training-free approaches for removing such language biases, and show this significantly improves performance on challenging benchmarks that fall into Scenario 2.}
    \label{fig:prior_difference}
\end{figure*}

\section{Introduction}
Vision-language models (VLMs) trained on web-scale datasets will likely serve as the foundation for next-generation visual understanding systems. One reason for their widespread adoption is their ability to be used in an ``off-the-shelf" (OTS) or zero-shot manner without fine-tuning for specific target applications. In this study, we explore their OTS use on the task of image-text retrieval (e.g., given an image, predict the correct caption out of $K$ options) across a suite of nine popular benchmarks. %

{\bf Challenges.} While the performance of foundational VLMs is impressive, many open challenges remain. Recent analyses~\cite{kamath2023text, aro} point out that leading VLMs such as CLIP~\cite{clip} may often degrade to ``bag-of-words" that confuse captions such as ``{\tt the horse is eating the grass}'' and ``{\tt the grass is eating the horse}''. This makes it difficult to use VLMs to capture {\em compositions} of objects, attributes, and their relations. But somewhat interestingly, large-scale language models (LLMs) trained for autoregressive next-token prediction~\cite{gpt3} seem to be able to discern such distinctions, which we investigate below. A related but under-appreciated difficulty is that of {\em benchmarking} the performance of visio-linguistic reasoning. 
Perhaps the most well-known example in the community is that of the influential VQA benchmarks~\cite{vqa}, which could be largely solved by exploiting linguistic biases in the dataset -- concretely, questions about images could often be answered by ``blind" language-only models that did not look at the image~\cite{vqa2}. Notably, we find that such blind algorithms still excel on many contemporary image-text retrieval benchmarks where VLMs may struggle.

{\bf Generative models for discriminative tasks.} We tackle the above challenges by revisiting the role of language priors through a probabilistic lens. To allow for a probabilistic treatment, we focus on generative VLMs that take an image as input and stochastically generate text via next-token prediction~\cite{blip, blip2}. We first demonstrate that such models can be easily repurposed for discriminative tasks (such as retrieval) by setting the match score for an image-text pair to be the probability that the VLM would generate that text from the given image, or $P(\text{text}|\text{image})$. We call this probability score the Visual Generative Pre-Training Score, or VisualGPTScore. Computing the VisualGPTScore is even more efficient than next-token generation since given an image, all tokens from a candidate text string can be evaluated in parallel. Though conceptually straightforward, such an approach is not a common baseline. In fact, the generative VLMs~\cite{blip} that we analyze train {\em separate} discriminative heads for matching/classifying image-text pairs, but we find that their language generation head itself produces better scores for matching (since it appears to better capture compositions). Indeed, the OTS VisualGPTScore performs surprisingly well on many benchmarks, even producing near-perfect accuracy on ARO~\cite{aro}. But it still struggles on other benchmarks such as Winoground~\cite{winoground}. We analyze this below. 

{\bf The role of language priors.} We analyze the discrepancy in performance across benchmarks from a probabilistic perspective. Our key insight is that many benchmark biases can be formalized as mismatching distributions over text between foundational pre-training data and benchmark test data -- $P_{train}(\text{text})$ versus $P_{test}(\text{text})$. We use a first-principles analysis to account for distribution shift by simply reweighting the VisualGPTScore with the Bayes factor $P_{test}(\text{text})/P_{train}(\text{text})$, a process we call {\em debiasing.} To compute the Bayes reweighting factor, we need access to both the train and test language prior. We compute $P_{train}(\text{text})$ from an OTS VLM by drawing Monte-Carlo samples of $P_{train}(\text{text}|\text{image})$ from the trainset or Gaussian noise images. Because $P_{test}(\text{text})$ may require access to the test set, we explore practical variants that assume $P_{test}$ is (a) identical to $P_{train}(\text{text})$, (b) uninformative/uniform, or (c) learnable from a small held-out valset. Our analysis helps explain the strong performance of the VisualGPTScore on certain benchmarks and its poor performance on others. Moreover, our analysis offers simple strategies to improve performance through debiasing without requiring any retraining. We conclude by showing a theoretical connection between debiasing and mutual information, which can be seen as a method for removing the effect of marginal priors when computing joint probability scores.

{\bf Empirical analysis.} We conduct a thorough empirical evaluation of the OTS VisualGPTScore (and its debiased variants) for open-sourced image-conditioned language models~\cite{blip, blip2, llava} across nine popular vision-language benchmarks. We first point out that the VisualGPTScore by itself produces SOTA accuracy on certain benchmarks like ARO~\cite{aro} where their inherent language biases help remove incorrect captions that are also unnatural (such as ``{\tt a white duck the its wings while in water}'' as shown in Fig.~\ref{fig:prior_difference}). In fact, we show that blind baselines also do quite well on these benchmarks, since language-only models can easily identify such implausible captions. However, such language biases do not work well on benchmarks where incorrect captions are carefully constructed to be realistic. Here, VisualGPTScore should be debiased so as not to naively prefer more common captions that score well under its language prior. %
Debiasing consistently improves performance on benchmarks such as Flickr30K~\cite{flickr30k} and Winoground~\cite{winoground}. Interestingly, we find that debiasing can also improve accuracy on the {\em train} set used to learn the generative VLMs, indicating that such models learn biased estimates of the true conditional distribution $P_{train}(\text{text}|\text{image})$. We describe this further in our \autoref{app:biased}. 
Finally, our approach sets a new state-of-the-art on image-text alignment~\cite{winoground, eqben}, showing potential to replace the widely-used CLIPScore~\cite{hessel2021clipscore} in text-to-image evaluation. In fact, our latest work~\cite{lin2024evaluating, li2024evaluating} extends VisualGPTScore to more powerful vision-language models trained on visual-question-answering (VQA) data, achieving further improvements.

{\bf Contributions:}
\begin{itemize}[topsep=0pt, partopsep=0pt, leftmargin=0.3in, itemsep=0.1pt, parsep=0.1pt]
\item We introduce VisualGPTScore to repurpose generative VLMs for discriminative (image-text retrieval) tasks.
\item Our analysis shows that language priors play a key role in addressing train-test distribution shifts, leading to a zero-shot debiasing technique that significantly improves performance on challenging benchmarks.
\item We find that many recent benchmarks for foundational VLMs like ARO can be largely solved by blind solutions (e.g., P(text)) that ignore images. This underscores the need to reevaluate language priors in vision-language benchmarks.
\end{itemize}

\section{Related works}

{\bf Vision-language models.} State-of-the-art VLMs like CLIP~\cite{clip} are pre-trained on web-scale image-text datasets~\cite{laion5b} using discriminative objectives like image-text contrastive (ITC)~\cite{clip} and image-text matching (ITM)~\cite{albef} loss, typically formulated as $P(\text{match}|\text{image},\text{text})$. These pre-trained models exhibit robust zero-shot and few-shot~\cite{lin2023multimodality, wortsman2022robust} performance on traditional discriminative tasks~\cite{deng2009imagenet, coco}, often on par with fully-supervised models. More recently, image-conditioned language models like Flamingo~\cite{flamingo} and BLIP~\cite{blip, blip2} incorporate generative objectives primarily for downstream tasks such as captioning~\cite{nocaps} and VQA~\cite{vqa2}.

{\bf Visio-linguistic compositionality.} Benchmarks like ARO~\cite{aro}, Crepe~\cite{crepe}, Winoground~\cite{winoground}, EqBen~\cite{eqben}, VL-CheckList~\cite{vlchecklist}, and SugarCrepe~\cite{sugarcrepe} show that discriminative scores of VLMs, such as ITCScore and ITMScore, fail on their image-text retrieval tasks that assess compositional reasoning. Concurrently, advances on these tasks often involve fine-tuning discriminative VLMs with more data. One of the most popular approaches, NegCLIP~\cite{aro}, augments CLIP using programmatically generated negatives from original texts. Extending this, subsequent studies propose more expensive and heavily-engineered solutions. SyViC~\cite{cascante2023going} fine-tunes VLMs on million-scale synthetic images to augment spatial, attributive, and relation understanding. SGVL~\cite{herzig2023incorporating} and Structure-CLIP~\cite{huang2023structure} sample negatives using costly scene graph annotations. MosaiCLIP~\cite{singh2023coarse} and SVLC~\cite{doveh2022teaching} use linguistic tools such as scene graph parsers and LLMs to design better negative captions. The most recent DAC~\cite{doveh2023dense} leverages a combination of foundation models including BLIP2, ChatGPT, and SAM to rewrite and augment image captions. In contrast, we demonstrate that OTS generative scores can outperform these costly approaches on compositionality benchmarks.

{\bf Generative pre-training and scoring. } Vision models trained with {\em discriminative} objectives often lack incentives to learn structure information~\cite{brendel2019approximating, tejankar2021fistful}. Similarly, early LLMs trained with {\em discriminative} approaches, such as BERT~\cite{bert} and RoBERTa~\cite{roberta}, have also been criticized as bag-of-words models insensitive to word order~\cite{bertolini2022testing, hessel2021effective, papadimitriou2022classifying, sinha2021masked}. Conversely, generative pre-trained LLMs~\cite{gpt2} demonstrate exceptional compositional understanding while pre-trained solely with a next-token prediction~\cite{bengio2003neural} loss. Furthermore, generative scores of LLMs~\cite{gpt4, flan, opt} have flexible usage in downstream tasks, such as text evaluation~\cite{bartscore, fu2023gptscore} and reranking~\cite{keskar2019ctrl}. While generative scores from VLMs have been previously used for discriminative tasks~\cite{tschannen2023image, miech2021thinking}, our work uniquely investigates the critical role of language priors and introduces the first debiasing solution that improves retrieval without the need for retraining. %

\begin{figure*}[t]
  \centering
  \scalebox{0.96}{
      \begin{tabular}{c|c}
          {\includegraphics[width=0.5\textwidth]{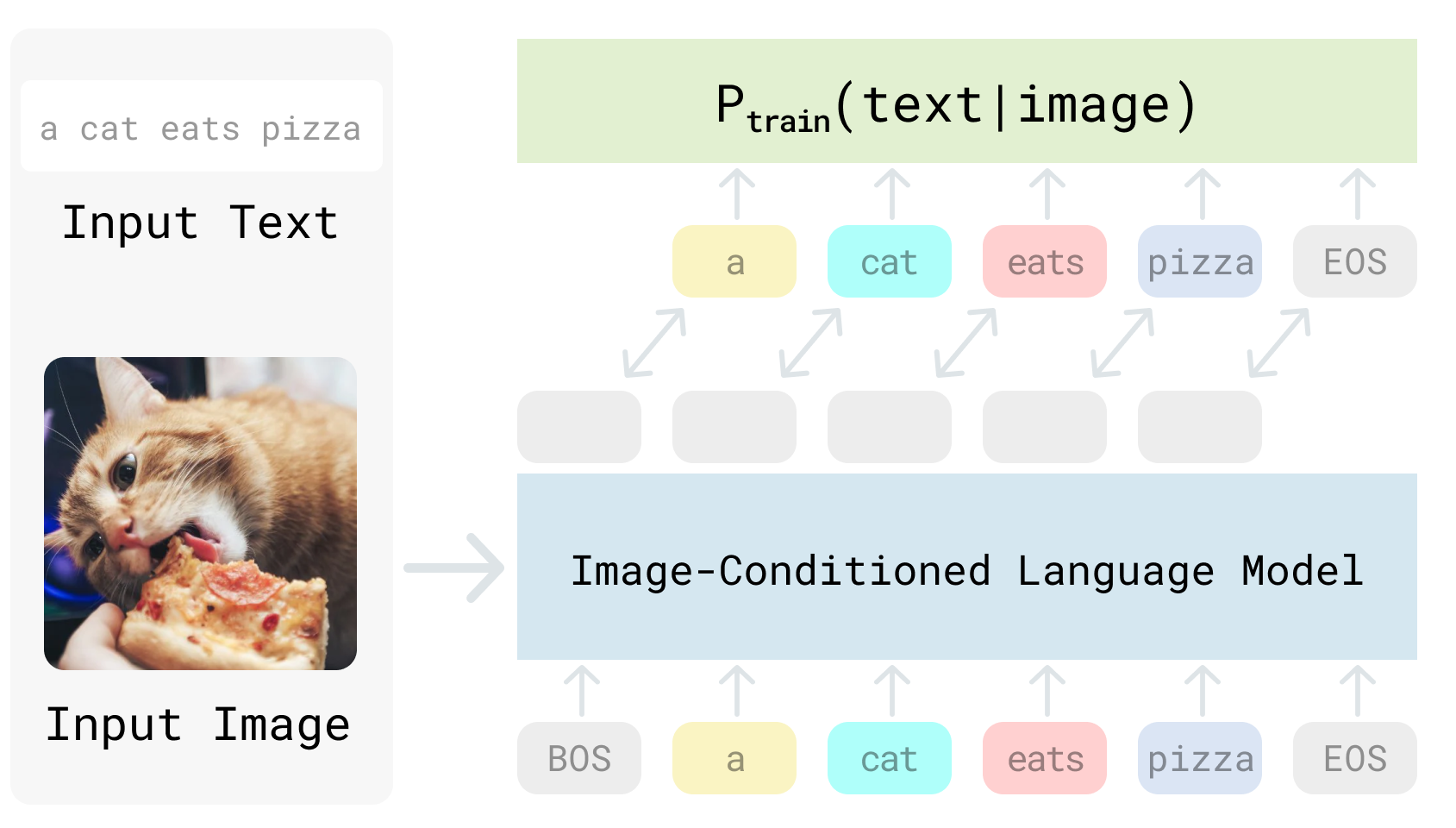}} & \includegraphics[width=0.45\textwidth]{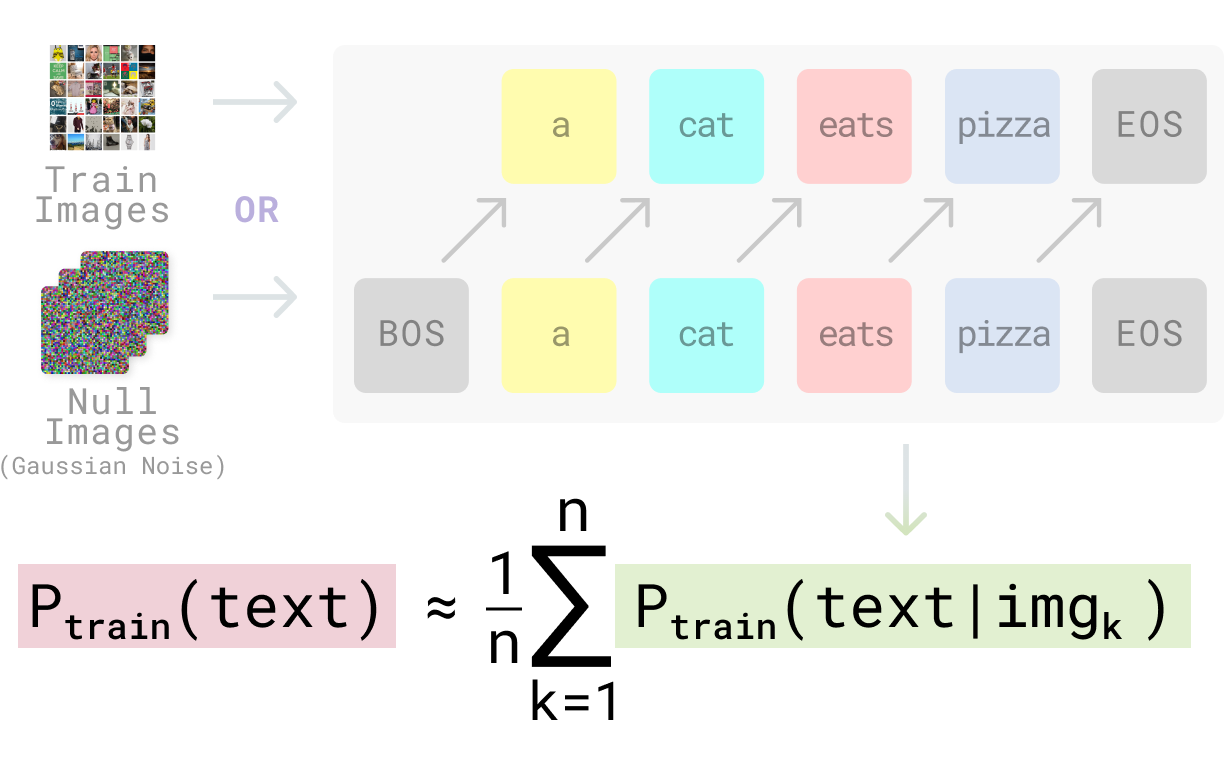} \\
          \multicolumn{1}{c}{(a) $P_{train}(\textbf{t}|\textbf{i})$ through generative VLMs} & \multicolumn{1}{c}{(b) $P_{train}(\textbf{t})$ via Monte Carlo sampling}
      \end{tabular}
  }
  \caption{\small {\bf Estimating $P_{train}(\mathbf{t}|\mathbf{i})$ and $P_{train}(\mathbf{t})$ from generative VLMs.} Figure (a) shows how image-conditioned language models such as \citet{blip} that generate text based on an image can be repurposed for computing $P_{train}(\mathbf{t}|\mathbf{i})$, which is factorized as a product of $\prod_{k=1}^m P(t_k | t_{<k}, \mathbf{i})$ for a sequence of $m$ tokens. These terms can be efficiently computed in {\em parallel}, unlike {\em sequential} token-by-token prediction for text generation. Figure (b) shows two approaches for Monte Carlo sampling of $P_{train}(\mathbf{t})$. While the straightforward approach is to sample trainset images, we find that using ``null'' (Gaussian noise) images can also achieve robust estimates.}
  \label{fig:visual_gpt_score}
\end{figure*}

\section{The role of language priors}
\label{sec:language_prior}

In this section, we present a simple probabilistic treatment for analyzing the role of language priors in image-conditioned language models (or generative VLMs). Motivated by their strong but inconsistent performance across a variety of image-text retrieval benchmarks, we analyze their behavior when there exists a mismatch between training and test distributions, deriving simple schemes for addressing the mismatch with reweighting. We emphasize that the training data that we refer to is the foundational pre-training dataset, while the test data is always a given benchmark dataset; in fact, most benchmarks we analyze do not even provide a trainset. %
We conclude by exposing a connection to related work on mutual information.

{\bf Computing $P(\mathbf{t}|{\bf i})$.} To begin our probabilistic treatment, we first show that image-conditioned language models (that probabilistically generate text based on an image) can be repurposed for computing a score between a given image ${\bf i}$ and text caption ${\bf t}$. The likelihood of a text sequence ${\mathbf t} = \{t_1, t_2, \cdots, t_m\}$ conditioned on image ${\mathbf i}$ is naturally factorized as an autoregressive product~\cite{bengio2003neural}:
\begin{align}
    P(\mathbf{t} | \mathbf{i}) = \prod_{k=1}^m P(t_k | t_{<k}, \mathbf{i})
\label{eq:product_of_conditionals}
\end{align}
Image-conditioned language models return back $m$ softmax distributions corresponding to the $m$ terms in the above expression. Text generation requires {\em sequential} token-by-token prediction, since token $t_k$ must be generated before it can be used as an input to generate the softmax distribution over token $t_{k+1}$. Interestingly, given an image ${\bf i}$ and a text sequence ${\bf t}$, the above probability can be computed in {\em parallel} because the entire sequence of tokens $\{t_k\}$ is already available as input. \autoref{fig:visual_gpt_score}-a shows a visual illustration.

{\bf Train-test shifts.} Given the image-conditioned model of $P({\bf t}|{\bf i})$ above, we now analyze its behavior when applied to %
test data distributions that differ from the %
trainset, denoted as $P_{test}$ versus $P_{train}$. 
Recall that any joint distribution over images and text can be factored into a product over a language prior and an image likelihood $P({\bf t},{\bf i}) = P({\bf t})P({\bf i}|{\bf t})$. Our analysis makes the strong assumption that the image likelihood $P(\mathbf{i}|\mathbf{t})$ is identical across the train and test data, but the language prior $P({\bf t})$ may differ. Intuitively, this assumes that the visual appearance of entities (such as a {\tt "white duck"}) remains consistent across the training and test data, but the frequency of those entities (as manifested in the set of captions $P({\bf t})$) may vary. %
We can now derive $P_{test}(\mathbf{t} | \mathbf{i})$ via Bayes rule:
\begin{align}
    P_{test}(\mathbf{t} | \mathbf{i}) &\propto P(\mathbf{i}|\mathbf{t}) P_{test}(\mathbf{t}) \\
    &= P(\mathbf{i}|\mathbf{t}) \frac{P_{train}(\mathbf{t})}{P_{train}(\mathbf{t})} P_{test}(\mathbf{t}) \\
    &\propto P_{train}(\mathbf{t}|\mathbf{i}) \frac{P_{test}(\mathbf{t})}{P_{train}(\mathbf{t})} 
\label{eq:optimal_i2t}
\end{align}
The above shows that the generative pre-training score $P_{train}(\mathbf{t}|\mathbf{i})$ need simply be weighted by the {\em ratio} of the language priors in the testset versus trainset. Intuitively, if a particular text caption appears {\em more} often in the testset than the trainset, one should {\em increase} the score reported by the generative model. However, one often does not have access to the text distribution on the testset. For example, real-world deployments and benchmark protocols may not reveal this. In such cases, one can make two practical assumptions; either the language distribution on test is identical to train, or it is uninformative/uniform (see \autoref{fig:prior_difference}):
\begin{align}
\text{Scenario 1:} \notag\\
\quad P_{test}(\mathbf{t}) &=  P_{train}(\mathbf{t}) &&\Rightarrow  &&\text{Optimal score is } P_{train}(\mathbf{t} | \mathbf{i}) \label{eq:same_prior}  \\
\text{Scenario 2:} \notag\\
\quad P_{test}(\mathbf{t}) &\text{ is uniform.} &&\Rightarrow &&\text{Optimal score is } \frac{P_{train}(\mathbf{t}|\mathbf{i})}{P_{train}(\mathbf{t})} \label{eq:uniform_prior}
\end{align}
{\bf Tunable $\alpha$.} In reality, a testset might be a mix of both scenarios. To model this, we consider a soft combination where the language prior on the testset is assumed to be a flattened version of the language prior on the trainset, for some temperature parameter $\alpha \in [0,1]$:
\begin{align}
P_{test}(\mathbf{t}) \propto P_{train}(\mathbf{t})^{1-\alpha} \quad \Rightarrow \text{Optimal score is }& \frac{P_{train}(\mathbf{t}| \mathbf{i})}{P_{train}({\bf t})^{\alpha}}
\label{eq:debias_i2t}
\end{align}
By setting $\alpha$ to 0 or 1, one can obtain the two scenarios described above. Some deployments
(or benchmarks) may benefit from tuning $\alpha$ on a held-out valset, if available.

{\bf Implications for retrieval benchmarks.} We speculate some benchmarks like ARO-Flickr~\cite{aro} are close to Scenario 1 because they include negative captions  that are {\em implausible}, such as ``{\tt a white duck the its wings while in water spreads}''. Such captions will have a low score under the language prior $P_{train}(\mathbf{t})$ and so reporting the raw generative score $P_{train}(\mathbf{t}|\mathbf{i})$ (that keeps its language prior or bias) will improve accuracy. In fact, we show that applying a {\em blind} language model (that ignores all image evidence) can itself often identify the correct caption. On the other hand, for test datasets with more {\em realistic} negative captions (Scenario 2), it may be useful to remove the language bias of the trainset, since that will prefer to match to common captions (even if they do not necessarily agree with the input image). This appears to be the case for SugarCrepe~\cite{sugarcrepe}, which uses LLMs like ChatGPT to ensure that the negative captions are realistic.

{\bf An information-theoretic derivation of $\alpha$-debiasing.} Our approach to debiasing is reminiscent of mutual information, which can also be seen as a method for removing the
effect of marginal priors when computing joint probability scores~\cite{daille1994approche}. In fact, $\alpha$-debiasing (Eq.~\ref{eq:debias_i2t}) is equivalent to a form of pointwise mutual information (PMI) known as PMI$^k$~\cite{role2011handling}. PMI is a classic information-theoretic measure that quantifies the association between two variables~\cite{yao2010co, henning2017estimating, shrivastava2021clip}. In the context of image-text retrieval, PMI measures how much more or less likely the image-text pair co-occurs than if the two were independent:
\begin{align}
    \text{pmi}_{P}(\mathbf{t}, \mathbf{i}) = \frac{P(\mathbf{t}, \mathbf{i})}{P(\mathbf{t}) P(\mathbf{i})} = \frac{P(\mathbf{i} | \mathbf{t})}{P(\mathbf{i})} = \frac{P(\mathbf{t} | \mathbf{i})}{P(\mathbf{t})} 
\label{eq:PMI}
\end{align}

However, directly applying PMI (Eq.~\ref{eq:PMI}) for retrieval tends to overly inflate scores for rarer texts~\cite{role2011handling}. Consequently, the PMI$^k$ approach was introduced to control the strength of debiasing. Below, we rewrite the Eq.~\ref{eq:debias_i2t} using the language of PMI$^k$:
\begin{align}
    \frac{P_{train}(\mathbf{t}| \mathbf{i})}{P_{train}({\bf t})^{\alpha}}
    &= \frac{P_{train}(\mathbf{t}, \mathbf{i})}{P_{train}(\mathbf{i}) P_{train}(\mathbf{t})^{\alpha}} \\
    &\propto \frac{P_{train}(\mathbf{t}, \mathbf{i})^{\frac{1}{\alpha}}}{P_{train}(\mathbf{i}) P_{train}(\mathbf{t})} \nonumber\\
    &\quad \quad \text{, as }P_{train}(\mathbf{i})\text{ is constant in I-to-T} \\
    &= \text{pmi}^{k}_{P_{train}}(\mathbf{t}, \mathbf{i}) \quad \text{, where } k = \frac{1}{\alpha} \geq 1 
\label{eq:alpha_pmi}
\end{align}
Eq.~\ref{eq:alpha_pmi} shows that our $\alpha$-debiasing is equivalent to PMI$^k$ for $k = \frac{1}{\alpha}$. PMI$^k$ is widely adopted in information retrieval tasks~\cite{pmimt1, pmimt2, pmiretrieval}. This alternative derivation could explain why $\alpha$-debiasing remains effective across various testing benchmarks (as we show next), even when our previous probabilistic assumptions may not hold.

\section{Experimental results on I-to-T retrieval}
\label{sec:visual_gpt_score}
In this section, we verify our hypothesis on I-to-T retrieval benchmarks using state-of-the-art multimodal generative VLMs. In particular, we adopt image-conditioned language models such as BLIP~\cite{blip} as the learned estimator of $P_{train}(\mathbf{t}|\mathbf{i})$. Then, we discuss how we perform Monte Carlo estimation of $P_{train}(\mathbf{t})$, including a novel efficient sampling method based on ``content-free'' Gaussian noise images. Finally, we show the state-of-the-art results of our generative approach on recent I-to-T retrieval benchmarks.

{\bf Preliminaries.} We leverage OTS image-conditioned language models to estimate $P_{train}(\mathbf{t})$. Most of our diagnostic experiments focus on the open-sourced BLIP~\cite{blip, blip2} model, trained on public image-text corpora using discriminative (ITC and ITM) and generative (captioning) objectives. Discriminative objectives typically model $P(\text{match}|\mathbf{t}, \mathbf{i})$. For example, ITCScore calculates cosine similarity scores between image and text features using a dual-encoder; ITMScore jointly embeds image-text pairs via a fusion-encoder and returns softmax scores from a binary classifier. We term the generative score as {\bf Visual} {\bf G}enerative {\bf P}re-{\bf T}raining Score ({\bf VisualGPTScore}). While BLIP is pre-trained using all three objectives, this generative score has not been applied to discriminative tasks before our work. Lastly, our approach can be extended to other generative VLMs. We also present some additional results using LLaVA-1.5~\cite{llava}, a recent state-of-the-art VLM~\cite{llava} that produces SOTA accuracy on several challenging benchmarks.

{\bf Implementing VisualGPTScore.} Our method calculates an average of the log-likelihoods of $t_k$ at each token position $k$ and applies an exponent to cancel the log:
\begin{align}
    \textbf{VisualGPTScore}(\mathbf{t}, \mathbf{i}) := e^{\frac{1}{m} \sum_{k=1}^m \log(P(t_k | t_{<k}, \mathbf{i}))}
\label{eq:visual_gpt_score}
\end{align}
To condition on an input image, BLIP uses a multimodal casual self-attention mask~\cite{blip} in its image-grounded text decoder, i.e., each text token attends to all its preceding vision and text tokens. We emphasize that VisualGPTScore has the same computational cost as ITMScore, which uses the same underlying transformer but with a bi-directional self-attention mask to encode an image-text pair. We address potential biases of this estimator in \autoref{app:biased}.

\begin{table*}[h!]
    \centering
    \caption{\small {\bf OTS generative VLMs are SOTA on image-to-text retrieval benchmarks.} We begin by evaluating \colorbox{softred}{blind language models (in red)}. Surprisingly, this already produces SOTA accuracy on certain benchmarks such as ARO-Flickr, compared to the \colorbox{softgray}{best discriminative approaches (in gray)}. We also find that blind inference of generative VLMs, \colorbox{softblue}{\smash{$P_{train}(\mathbf{t})$ via sampling Gaussian noise images (in blue)}\vphantom{g}}, often performs better and achieve above-chance performance even on the most recent SugarCrepe. Next, we show that simply repurposing a generative VLM's language generation head for computing image-text scores \colorbox{softyellow}{(VisualGPTScore in yellow)}, which corresponds to $\alpha = 0$, consistently produces SOTA accuracy across all benchmarks. Finally, debiasing this score by \colorbox{softgreen}{tuning $\alpha$ on valset (in green)} further improves performance, establishing the new SOTA.} 
    \scalebox{0.8}{
    \begin{tabular}{cc}
        \begin{NiceTabular}{llcccc}
        \CodeBefore
          \rectanglecolor{softred}{6-3}{8-6}
          \rectanglecolor{softblue}{9-3}{9-6}
          \rectanglecolor{softyellow}{22-3}{22-6}
          \rectanglecolor{softgreen}{24-3}{24-6}
        \Body
        \toprule
               \multirow{2}{*}{Score} &  \multirow{2}{*}{Method} & \multicolumn{4}{c}{\bf ARO} 
               \\ \cmidrule{3-6}
               & & Rel&  Attr&  COCO& Flickr\\ 
             \midrule
               Random & - &  50.0&  50.0&  20.0& 20.0\\ \hline
               \multirow{2}{*}{Text-Only} & Vera & 61.7 & 82.6 & 59.8 & 63.5 \\ 
               & Grammar & 59.6 & 58.4 & 74.3 & 76.3  \\ \hline
               \multirow{3}{*}{$P_{LLM}(\mathbf{t})$} & BART & 81.1& 73.6 & 95.0 & 95.2 \\
               & Flan-T5 & 84.4& 76.5 & 98.0 & 98.2 \\
               & OPT & 84.7 & 79.8 & 97.9 & 98.6
               \\  \hline
               $P_{train}(\mathbf{t})$ & BLIP & 87.6 &	80.7 &	98.6 &	99.1 \\ \hline
               \multirow{13}{*}{$P(\text{match}|\mathbf{t},\mathbf{i})$}&  CLIP &  59.0 &  62.0 &  46.0 & 60.0 \\ 
               &  LAION2B-CLIP &   51.6 & 61.9 & 25.2 & 30.2 \\ 
               &  LAION5B-CLIP &   46.1 & 57.8 & 26.1 & 31.0 \\ 
               &  NegCLIP &  81.0 &  71.0 &  86.0 & 91.0 \\ 
               &  Structure-CLIP &  \cellcolor{softgray}83.5 &  85.1 &  - & - \\ 
               &  SyViC &  80.8 & 72.4 &  92.4 & 87.2 \\ 
               &  SGVL &  - &  - & 87.2 & 91.0 \\ 
                &  MosaiCLIP &  82.6 &  78.0 &  87.9 & 86.3 \\ 
      & DAC-LLM & 81.3 & 73.9 & \cellcolor{softgray}94.5 & \cellcolor{softgray}95.7 \\ 
      & DAC-SAM & 77.2 & 70.5 & 91.2 & 93.9 \\ 
      & BLIP-ITC & 63.1 & 81.6 &	34.3 &	41.7 \\
      & BLIP-ITM & 58.7 &	\cellcolor{softgray}90.3 &	45.1 & 51.3 \\ \hline
      \multirow{3}{*}{\Large $\frac{P_{train}(\mathbf{t}|\mathbf{i})}{P_{train}(\mathbf{t})^\alpha}$} & Ours ($\alpha = 0$) & {\bf 89.1} &	{\bf 95.3} &	{\bf 99.4} &	{\bf 99.5} \\
      & Ours ($\alpha = 1$) & 68.1 &	87.9 &	32.4 &	44.5 \\
      & Ours ($\alpha = \alpha^\ast$) & {\bf 89.1} &  {\bf 95.4} &	{\bf 99.4} &	{\bf 99.5} \\
     \bottomrule
        \end{NiceTabular}
        & \quad
        
        \begin{NiceTabular}{llccc}
        \CodeBefore
          \rectanglecolor{softred}{6-3}{8-5}
          \rectanglecolor{softblue}{9-3}{9-5}
          \rectanglecolor{softyellow}{21-3}{21-5}
          \rectanglecolor{softgreen}{23-3}{23-5}
        \Body
        
        \toprule
               \multirow{2}{*}{Score} &  \multirow{2}{*}{Method} & \multicolumn{3}{c}{\bf VL-CheckList} 
               \\ \cmidrule{3-5}
               & & Object&  Attribute&  Relation\\ 
             \midrule
               Random & - &  50.0&  50.0&  50.0\\ \hline
               \multirow{2}{*}{Text-Only} & Vera & 82.5 & 74.0 & 85.7 \\ 
               & Grammar & 58.0 & 52.4 & 68.5  \\ \hline
               \multirow{3}{*}{$P_{LLM}(\mathbf{t})$} & BART & 52.0& 51.0 & 45.1 \\
               & Flan-T5 & 60.3& 55.0 & 49.3 \\
               & OPT & 59.3 & 48.8 & 60.0 
               \\  \hline
               $P_{train}(\mathbf{t})$ & BLIP & 68.2 &	58.7 &	75.9 \\ \hline
               \multirow{9}{*}{$P(\text{match}|\mathbf{t},\mathbf{i})$}&  CLIP &  81.6 &  67.6 &  63.1 \\ 
               &  LAION2B-CLIP &   84.7 & 67.8 & 66.5 \\
               &  LAION5B-CLIP &   87.9 & 70.3 & 63.9 \\
               &  NegCLIP &  81.4 &  72.2 &  63.5 \\ 
               &  SyViC &  - & 70.4 &  69.4 \\ 
               &  SGVL &  85.2 &  78.2 & 80.4 \\ 
                &  SLVC &  85.0 &  72.0 &  69.0 \\ 
                & DAC-LLM & 87.3 & 77.3 & 86.4 \\ 
      & DAC-SAM & 88.5 & 75.8 & \cellcolor{softgray}89.8 \\ 
      & BLIP-ITC & \cellcolor{softgray}90.6 & 80.3 &	73.5 \\
      & BLIP-ITM & 89.9 &	\cellcolor{softgray}80.7 & 67.7	 \\ \hline
      \multirow{3}{*}{\Large $\frac{P_{train}(\mathbf{t}|\mathbf{i})}{P_{train}(\mathbf{t})^\alpha}$} & Ours ($\alpha = 0$) & 92.6 & 78.7 & 90.8 \\
      & Ours ($\alpha = 1$) & 90.4 &	77.6   & 77.8 \\
      & Ours ($\alpha = \alpha^\ast$) & {\bf 94.4} &  {\bf 82.1} &	{\bf 92.8} \\
     \bottomrule
        \end{NiceTabular}
        
         \\ 
        \multicolumn{1}{c}{\Large \bf (a) Accuracy on ARO} & \multicolumn{1}{c}{\Large \bf (b) Accuracy on VL-CheckList} \\ \\
        \begin{NiceTabular}{llccc}
        \CodeBefore
          \rectanglecolor{softred}{6-3}{8-5}
          \rectanglecolor{softblue}{9-3}{9-5}
          \rectanglecolor{softyellow}{16-3}{16-5}
          \rectanglecolor{softgreen}{18-3}{18-5}
        \Body
        \toprule
                \multirow{2}{*}{Score} &  \multirow{2}{*}{Method} & \multicolumn{3}{c}{\bf SugarCrepe} 
               \\ \cmidrule{3-5}
                &   &  Replace &  Swap &  Add \\ 
             \midrule
             Random & - &  50.0&  50.0&  50.0\\ \hline
               \multirow{2}{*}{Text-Only} & Vera & 49.5 & 49.3 & 49.5 \\ 
               & Grammar & 50.0 & 50.0 & 50.0  
               \\  \hline
               \multirow{3}{*}{$P_{LLM}(\mathbf{t})$} & BART & 48.4& 51.9 & 61.2 \\
               & Flan-T5 & 51.4& 57.6 & 40.9  \\
               & OPT & 58.5 & 66.6 & 45.8
               \\  \hline
               $P_{train}(\mathbf{t})$ & BLIP & 75.9 &	77.1 &	70.9 \\ \hline
               \multirow{6}{*}{$P(\text{match}|\mathbf{t},\mathbf{i})$}&  CLIP &  80.8 &  63.3 &  75.1 \\ 
               &  LAION2B-CLIP &  86.5 &  68.6 &  88.4 \\ 
               &  LAION5B-CLIP &  85.0 &  68.0 &  89.6 \\
               &  NegCLIP &  88.3 &  76.2 &  \cellcolor{softgray}90.2 \\ 
              & BLIP-ITC & 85.8 & 73.8 &	85.7  \\
              & BLIP-ITM & \cellcolor{softgray}88.7 &	\cellcolor{softgray}81.3 &	87.6 \\ 
              \hline
              \multirow{3}{*}{\Large $\frac{P_{train}(\mathbf{t}|\mathbf{i})}{P_{train}(\mathbf{t})^\alpha}$} & Ours ($\alpha = 0$) & 93.3 &	91.0 &	91.0 \\
              & Ours ($\alpha = 1$) & 83.2 &	85.5 &	85.9\\
              & Ours ($\alpha = \alpha^\ast$) & {\bf 95.1} &  {\bf 92.4} &	{\bf 97.4}  \\
             \bottomrule
        \end{NiceTabular}
        & \quad
        
        \begin{NiceTabular}{llccc}
        \CodeBefore
          \rectanglecolor{softred}{6-3}{8-5}
          \rectanglecolor{softblue}{9-3}{9-5}
          \rectanglecolor{softyellow}{15-3}{15-5}
          \rectanglecolor{softgreen}{17-3}{17-5}
        \Body
        \toprule
                \multirow{2}{*}{Score} &  \multirow{2}{*}{Method} & \multicolumn{3}{c}{\bf Crepe} 
               \\ \cmidrule{3-5}
                &   &  Atom &  Swap &  Negate \\ 
             \midrule
             Random & - &  16.7&  16.7&  16.7\\ \hline
               \multirow{2}{*}{Text-Only} & Vera & 43.7 & 70.8 & 66.2 \\ 
               & Grammar & 18.2 & 50.9 & 9.8  
               \\  \hline
               \multirow{3}{*}{$P_{LLM}(\mathbf{t})$} & BART & 38.8 & 53.3 & 44.4  \\
               & Flan-T5 & 43.0& 69.5 & 13.6   \\
               & OPT & 53.3& 72.7 & 5.0 
               \\  \hline
               $P_{train}(\mathbf{t})$ & BLIP & 55.4 &	69.7 &	60.8  \\ \hline
               \multirow{5}{*}{$P(\text{match}|\mathbf{t},\mathbf{i})$}&  CLIP &  22.3 &  \cellcolor{softgray}26.6 &  \cellcolor{softgray}28.8 \\ 
               &  LAION2B-CLIP &  23.6 &  24.8 &  18.0 \\ 
               &  LAION5B-CLIP &  24.2 &  23.9 &  20.1 \\
               & BLIP-ITC & 24.8 & 17.7 &	26.5 \\ 
              & BLIP-ITM & \cellcolor{softgray}29.5  & 20.7 &	25.5  \\ 
              \hline
              \multirow{3}{*}{\Large $\frac{P_{train}(\mathbf{t}|\mathbf{i})}{P_{train}(\mathbf{t})^\alpha}$} & Ours ($\alpha = 0$) & {\bf 73.2} & {\bf 78.1} & {\bf 79.6} \\
              & Ours ($\alpha = 1$) & 20.6 &	28.3 &	35.6\\
              & Ours ($\alpha = \alpha^\ast$) & {\bf 73.3} &  {\bf 78.1} &	{\bf 79.6}  \\
             \bottomrule
        \end{NiceTabular}
         \\ 
        \multicolumn{1}{c}{\Large \bf (c) Accuracy on SugarCrepe} & \multicolumn{1}{c}{\Large \bf (d) Accuracy on Crepe} \\ 
    \end{tabular}
    }
    \label{tab:i_to_t_retrieval}
\end{table*}

{\bf Estimating $P_{train}(\mathbf{t})$ using Monte Carlo sampling (oracle approach).} Given $P_{train}(\mathbf{t}|\mathbf{i})$, we can estimate $P_{train}(\mathbf{t})$ via classic Monte Carlo sampling~\cite{shapiro2003monte}, by drawing $n$ images from the train distribution, such as LAION114M~\cite{laion400m} for BLIP:
\begin{align}
    P_{train}(\mathbf{t}) \approx \frac{1}{n} \sum_{k=1}^n P_{train}(\mathbf{t} | \mathbf{i}_k) %
\label{eq:monte_carlo}
\end{align}

{\bf Reducing sampling cost with Gaussian noise images (our approach).} The above \autoref{eq:monte_carlo} requires many trainset samples to achieve robust estimates. To address this, we draw inspiration from \cite{calibratellm}, which uses a {\em content-free} text prompt ``{\tt N/A}'' to calibrate the probability of a text from LLMs, i.e., $P(\mathbf{t}|\text{``{\tt N/A}''})$. To apply this to our generative VLMs, we choose to sample ``null'' inputs as Gaussian noise images. It turns out Eq.~\ref{eq:monte_carlo} can be estimated using as few as 1-3 Gaussian noise images (with a mean and standard deviation calculated from trainset distribution). We provide a visual illustration of this method in \autoref{fig:visual_gpt_score}-b. We find this method to be less computationally demanding and just as effective as sampling thousands of images from trainset. We ablate sampling procedures in \autoref{app:alpha_ablation} and show that our method generalizes across BLIP and BLIP-2 architectures in \autoref{app:blip_2}.

{\bf Benchmarks and evaluation protocols.} We comprehensively report on four recent I-to-T retrieval benchmarks that assess compositionality, including ARO~\cite{aro}, Crepe~\cite{crepe}, SugarCrepe~\cite{sugarcrepe}, and VL-CheckList~\cite{vlchecklist}. In these datasets, each image has a single positive caption and multiple negative captions. ARO~\cite{aro} has four datasets: VG-Relation, VG-Attribution, COCO-Order, and Flickr30k-Order. SugarCrepe~\cite{sugarcrepe} has three datasets: Replace, Swap, and Add. For Crepe~\cite{crepe}, we use the entire productivity set and report on three datasets: Atom, Negate, and Swap. VL-CheckList~\cite{vlchecklist} has three datasets: Object, Attribute, and Relation. \autoref{app:dataset_visualization} visualizes these datasets.

{\bf SOTA performance on all four benchmarks. } In \autoref{tab:i_to_t_retrieval}, we show that our OTS generative approaches, based on the BLIP model pre-trained on LAION-114M with ViT-L image encoder, achieves state-of-the-art results on all benchmarks. We outperform the best discriminative VLMs, including LAION5B-CLIP, and consistently surpass other heavily-engineered solutions, including NegCLIP, SyViC, MosaiCLIP, DAC, SVLC, SGVL, Structure-CLIP, all of which fine-tune CLIP on much more data. Details on how we report the baseline results can be found in \autoref{app:other}. For reference, we also include results of text-only Vera and Grammar from \citet{sugarcrepe}. To show that even the most recent SugarCrepe is not exempt from language biases, we run two more text-only methods:
\begin{enumerate}[topsep=0pt, partopsep=0pt, itemsep=2pt, parsep=2pt, leftmargin=0.3in]
    \item $P_{LLM}(\mathbf{t})$: passing captions into a pure LLM, such as BART-base~\cite{bartscore}, FLAN-T5-XL~\cite{flan}, and OPT-2.7B~\cite{opt}, to compute a text-only GPTScore~\cite{fu2023gptscore}.
    \item $P_{train}(\mathbf{t})$: passing both captions and Gaussian noise images to BLIP as shown in \autoref{fig:visual_gpt_score}.
\end{enumerate}

\begin{table*}[h!]
\centering
\caption{\small {\bf $\alpha$-debiasing on I-to-T benchmarks and $P_{train}(\mathbf{t})$ frequency charts of both positive and negative captions.} Increasing $\alpha$ from 0 to 1 hurts performance
on benchmarks with non-sensical negative captions like ARO-Flickr. ARO's negative captions are easier to identify because of their low score under the language prior $P_{train}(\mathbf{t})$, implying such benchmarks may even be solved with blind algorithms that avoid looking at images. On the other hand, for benchmarks like SugarCrepe with more balanced $P_{train}(\mathbf{t})$ between positive and negative captions, tuning $\alpha$ leads to performance gain. \autoref{app:other} shows analysis on all datasets.
}
\scalebox{0.9}{
\begin{tabular}{cc|cc}
\toprule
\textbf{Alpha-Tuning} & \textbf{Prior Frequency} & \textbf{Alpha-Tuning} & \textbf{Prior Frequency}  \\
\includegraphics[width=0.19\textwidth]{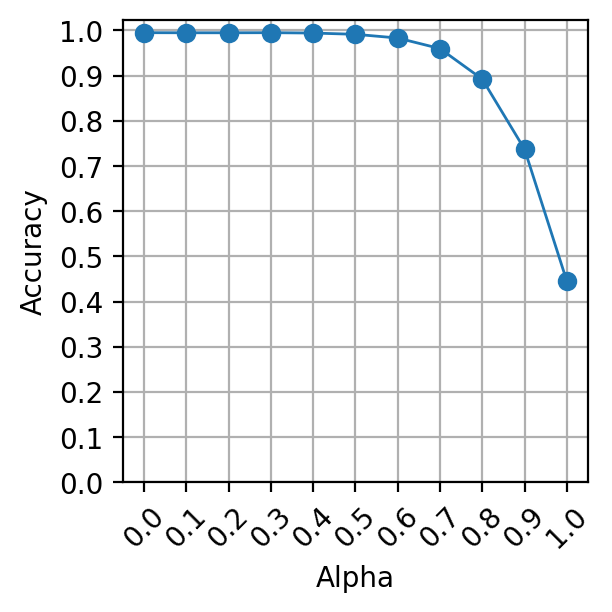} & \includegraphics[width=0.31\textwidth]{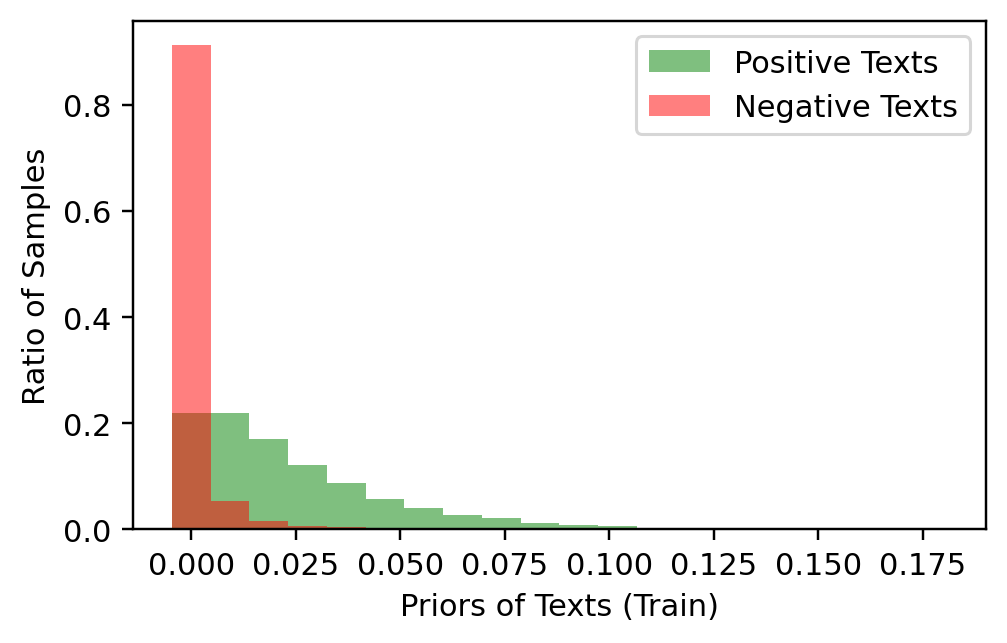} & \includegraphics[width=0.19\textwidth]{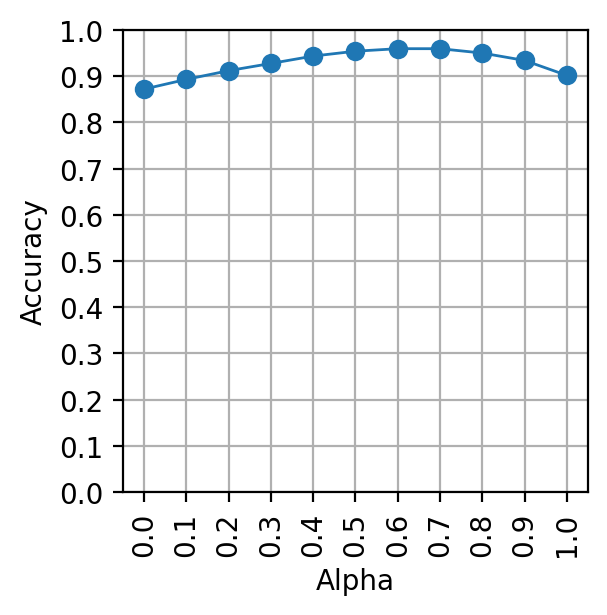} & \includegraphics[width=0.31\textwidth]{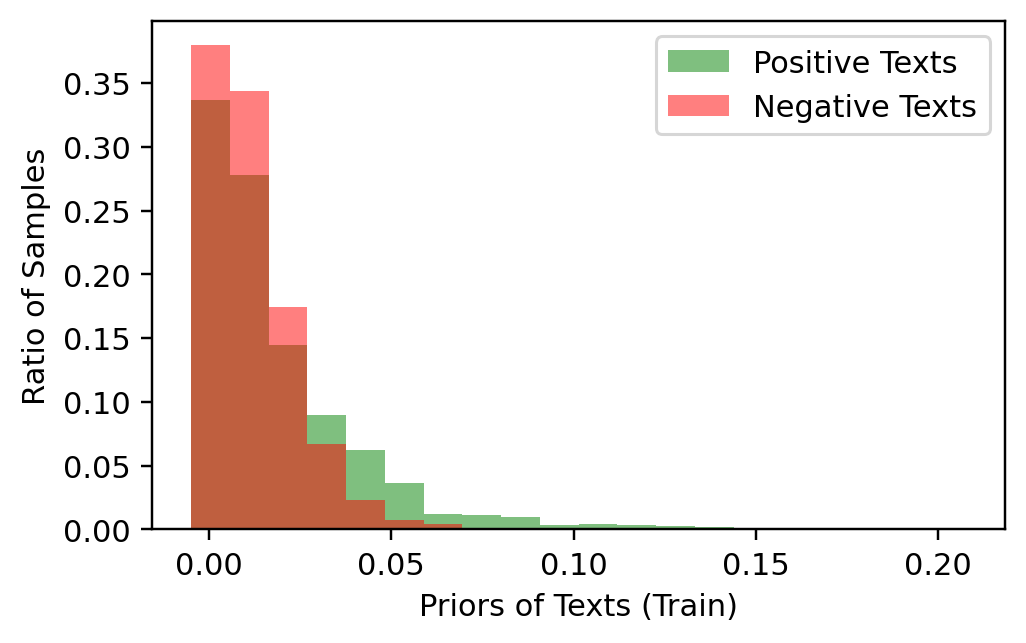}\\
\multicolumn{2}{c|}{ARO-Flickr} & \multicolumn{2}{c}{SugarCrepe-Add} \\
\bottomrule
\end{tabular}
}
\label{tab:datasets_plots_short}
\end{table*}

{\bf Discussion on $\alpha$-debiasing.} \autoref{tab:datasets_plots_short} shows that debiasing affects benchmarks differently depending on their construction; benchmarks with unrealistic negative captions (such as ARO-Flickr) benefit from a language prior that can identify such negative examples. Here, debiasing with large $\alpha$ hurts performance. On the other hand, benchmarks with realistic negative captions (such as SugarCrepe) tend to benefit from debiasing because it reduces the influence of the language prior.
Our findings are reminiscent of the lessons from the VQA benchmark~\cite{vqa2}, known to be solvable by ``blind'' algorithms that do not look at the image, e.g., questions such as  ``Is there a clock'' have an answer of ``Yes'' $98\%$ of the time. However, we also find that some recent benchmarks such as Winoground~\cite{winoground} and EqBen~\cite{eqben} introduce strict evaluation protocols that aggressively penalize such blind algorithms. We discuss these challenging Scenario 2 benchmarks (with far lower SOTA accuracy) in the next section. %

\section{Additional Challenging Benchmarks}
\label{sec:additional_exp}

\begin{table*}[h!]
  \centering
  \caption{\small {\bf Additional results on Winoground/EqBen/COCO/Flickr30K/ImageNet1K.} Table (a) shows the importance of $\alpha$-debiasing on these compositionality and large-scale retrieval benchmarks. While OTS generative scores do not work well, debiasing with a larger $\alpha$ close to 1 can consistently and often significantly improve I-to-T performance. To highlight the improvement, we mark \colorbox{softyellow}{results without debiasing ($\alpha=0$) (in yellow)}, \colorbox{softpink}{debiasing with a fixed $\alpha=1$ (in pink)}, and \colorbox{softgreen}{cross-validation using held-out valsets ($\alpha=\alpha_{val}^\ast$) (in green)}. Table (b) shows that OTS generative scores can obtain favorable results on all T-to-I retrieval tasks, competitive with the ITMScore. }
  \scalebox{0.74}{
    \begin{tabular}[h]{cc}
        \begin{NiceTabular}{llccccc}
            \CodeBefore
              \rectanglecolor{softyellow}{3-4}{7-4}
              \rectanglecolor{softpink}{3-5}{7-5}
              \rectanglecolor{softgreen}{3-6}{7-6}
            \Body
            \toprule
            \multirow{2}{*}{Metric} & \multirow{2}{*}{Benchmark}
            & \multirow{2}{*}{ITMScore} & \multicolumn{4}{c}{\Large $\frac{P_{train}(\mathbf{t}|\mathbf{i})}{P_{train}(\mathbf{t})^{\alpha}}$} \\ \cmidrule{4-7} 
            & & & $\alpha$=0 & $\alpha$=1 & $\alpha$=$\alpha^{\ast}_{val}$ & $\alpha^{\ast}_{val}$  \\
            \midrule
            \multirow{2}{*}{Text Score}& Winoground   & 35.5$_{(2.4)}$ & 27.5$_{(2.3)}$ & 33.7$_{(2.4)}$ & 36.6$_{(2.6)}$ &  0.855$_{(0.023)}$  \\
            & EqBen   & 26.1$_{(0.3)}$ & 9.6$_{(0.2)}$ & 19.8$_{(0.3)}$ & 19.8$_{(0.3)}$ &  0.992$_{(0.007)}$ \\
            \midrule
            \multirow{2}{*}{{\small R@1 / R@5}}& COCO   & 71.9 / 90.6 & 19.7 / 40.6 & 46.2 / 73.1 & 48.0 / 74.2 & 0.819 \\
            & Flickr30k   & 88.8 / 98.2 & 34.6 / 59.0 & 58.7 / 88.0 & 63.6 / 89.2 & 0.719 \\
            \midrule
            {\small Accuracy}& ImageNet1K   & 37.4 & 18.6 & 36.2 & 40.0 & 0.670 \\

            \bottomrule
        \end{NiceTabular}
        &
        \begin{tabular}{llcc}
            \toprule
            \multirow{1}{*}{Metric} & \multirow{1}{*}{Benchmark}
            & \multirow{1}{*}{{\footnotesize ITMScore}} & {$P_{train}(\mathbf{t}|\mathbf{i})$}\\
            \midrule
            \multirow{2}{*}{{Image Score}}& Winoground &  15.8 & 21.5\\
            & EqBen   & 20.3 & 26.1\\
            \midrule
            \multirow{2}{*}{{\footnotesize R@1 / R@5}}& COCO   &  54.8 / 79.0 & 55.6 / 79.2\\
            & Flickr30k   & 77.8 / 93.9 & 76.8 / 93.4\\

            \bottomrule
        \end{tabular} \\
        \multicolumn{1}{c}{{\large (a) $\alpha$-debiasing on valsets for I-to-T retrieval}} & \multicolumn{1}{c}{{\large (b) T-to-I retrieval}}
    \end{tabular}
    }
    \label{tab:other_benchmarks}
\end{table*}

In this section, we apply our OTS generative approaches to five more Scenario 2 benchmarks: (a) Winoground~\cite{winoground} and EqBen~\cite{eqben} for image-text alignment; (b) COCO~\cite{coco} and Flickr30K~\cite{flickr30k} for large-scale retrieval; (c) ImageNet~\cite{deng2009imagenet} for zero-shot image classification. While naively applying OTS VisualGPTScore leads to inferior performance on these benchmarks, our training-free $\alpha$-debiasing consistently improves its performance even with a fixed $\alpha$=1, without accessing the held-out valset (\autoref{tab:other_benchmarks}-a). We also derive the optimal text-to-image (T-to-I) retrieval objective and show that OTS generative scores can achieve robust T-to-I performance (\autoref{tab:other_benchmarks}-b). Lastly, we apply VisualGPTScore and its $\alpha$-debiased version to a state-of-the-art VLM, LLaVA-1.5~\cite{llava}, and outperform widely-used methods such as CLIPScore~\cite{hessel2021clipscore} on the challenging Winoground and EqBen benchmarks.
This suggests that VisualGPTScore is a superior choice for measuring image-text alignment.%

{\bf Balanced evaluation protocols for retrieval.} Winoground and EqBen evaluate image-text alignment through retrieval tasks, and we find their evaluation protocols discourage blind solutions. %
We refer the reader to the benchmarks for more details, but in summary, both benchmarks operate on pairs of image-text pairs $\{(\mathbf{i}_0, \mathbf{t}_0),(\mathbf{i}_1, \mathbf{t}_1)\}$ and construct two I-to-T retrieval (text score) tasks with a single image and two candidate captions. %
The text score is awarded 1 point only if {\em both} retrieval tasks are correct. Consider the common case where one caption is more likely under a language prior; here the common caption will be correctly retrieved for one of the tasks but will be incorrectly retrieved for the other, implying {\em no} points will be awarded. Similarly stringent metrics are used for T-to-I retrieval (image score). The final group score is awarded 1 point only if all 4 retrieval tasks are correct.

 \begin{table}[h]
\centering
\renewcommand{\arraystretch}{1.3}
\caption{\textbf{Superior performance of VisualGPTScore on challenging image-text alignment benchmarks.} We compare VisualGPTScore (and its $\alpha$=1 version) against popular image-text scoring methods such as CLIPScore and those that combine VLMs with additional LLMs like ChatGPT. On Winoground and EqBen, our \colorbox{softyellow}{VisualGPTScore ($\alpha$=0)} outperforms all methods using only a state-of-the-art VLM (LLaVA-1.5). Moreover, \colorbox{softpink}{debiasing with $\alpha$=1} (using a single Gaussian noise image) consistently improves I-to-T retrieval, thereby increasing the text and group score. To ensure a fair comparison, we use the publicly available model checkpoints and corresponding code of prior works. Method descriptions and implementation details can be found in \autoref{app:other}.}
\scalebox{0.6}{
\begin{NiceTabular}{lccccccccccc}
            \CodeBefore
              \rectanglecolor{softyellow}{12-3}{12-8}
              \rectanglecolor{softpink}{13-3}{13-8}
            \Body
\toprule
\multirow{2}{*}{\textbf{Method}} & \multirow{2}{*}{\textbf{LLMs used}} & \multicolumn{3}{c}{\textbf{Winoground}} &
\multicolumn{3}{c}{\textbf{EqBen}}
\\ 
\cmidrule(l){3-5} \cmidrule(l){6-8}
  &   &    Text & Image & Group & Text & Image  & Group  \\ \midrule
Random Chance & -- & 25.0 & 25.0 & 16.7 & 25.0 & 25.0 & 16.7 \\ \midrule
{\it Official implementation} \\
CLIPScore & -- & 31.3 & 11.0 & 8.8 & 35.0 & 33.6 & 21.4 \\
VPEval & ChatGPT & 12.8 &	11.0 &	6.3 &	34.3 &	25.7 &	21.4 \\
{LLMScore} & ChatGPT & 21.3 & 17.8 & 12.5 & 32.9 & 27.9 & 22.9 \\
\midrule
\multicolumn{6}{l}{\it Our results based on LLaVA-1.5} \\
\multirow{1}{*}{TIFA} & Llama-2 & 22.8 &	18.5 &	15.5 &	30.0 &	30.0 &	21.4 \\
\multirow{1}{*}{VQ2}  & FlanT5 & 14.0 &	27.3 &	10.0 &	22.9 &	40.7 &	20.0 \\
\multirow{1}{*}{Davidsonian} & ChatGPT &21.0 &	16.8 &	15.5 &	26.4 &	20.0 &	20.0 \\
\multirow{1}{*}{VisualGPTScore ($\alpha$=0)} & --   & 36.3 &	{\bf 37.0} & 24.8 &	25.7 &	{\bf 42.1} &	21.4 \\
\multirow{1}{*}{VisualGPTScore ($\alpha$=1)} & --   & {\bf 44.3} &	{\bf 37.0} &	{\bf 27.5} &	{\bf 42.9} &	{\bf 42.1} &	{\bf 29.3} \\
\bottomrule
\end{NiceTabular}
}
\label{tab:llava}
\end{table}

{\bf $\alpha$-debiasing consistently improves I-to-T retrieval.} \autoref{tab:other_benchmarks}-a shows that 
simply debiasing VisualGPTScore with a fixed $\alpha=1$ significantly improves performance on challenging I-to-T benchmarks. One can also do slightly better by using a held-out valset to tune for the optimal $\alpha \in [0, 1]$. For Winoground and EqBen, we sample half of the data as a valset and perform a grid search for $\alpha^{\ast}_{val}$ (using a step size of 0.001), reporting the performance on the other half. We repeat this process 10 times and report the mean and standard deviation. 
For COCO and Flickr30K, we perform $\alpha$-debiasing using Recall@1 (R@1) on the official valset. 
We report the zero-shot classification accuracy on ImageNet1K, which can be viewed as an I-to-T retrieval task that retrieves the best textual label (out of 1000) for each image. We simply use one-shot samples from \citet{lin2023multimodality} to cross validate on ImageNet, which incurs negligible costs. 
\autoref{app:alpha_ablation} details the debiasing procedure for each dataset.  
Lastly, we observe that generative approaches still lag behind the ITMScore of BLIP for the two large-scale retrieval benchmarks. This motivates us to study biases of generative models from the statistical perspective of biased estimators, briefly examined in \autoref{app:biased}. 

{\bf Extending to T-to-I retrieval.} Though not the focus of our work, we show that image-conditioned language models can be applied to T-to-I retrieval. Given a text caption $\mathbf{t}$, we can rewrite the Bayes optimal T-to-I retrieval objective as:
\begin{align}
    P_{test}(\mathbf{i}|\mathbf{t}) &\propto P_{train}(\mathbf{t}| \mathbf{i}) * P_{train}(\mathbf{i}) 
\label{eq:text_to_image}
\end{align}
\autoref{eq:text_to_image} is hard to implement because we do not have access to $P_{train}(\mathbf{i})$. However, when $P_{train}(\mathbf{i})$ is approximately uniform, one can directly apply $P_{train}(\mathbf{t}|\mathbf{i})$ for optimal performance. We report T-to-I performance in \autoref{tab:other_benchmarks}-b, where our generative approach obtains competitive results compared against ITMScore, likely because T-to-I retrieval is less affected by language biases.

{\bf State-of-the-art image-text alignment.} Text-to-image generative models %
such as DALL-E 3~\cite{dalle3} %
are often evaluated with models that score the agreement (or alignment) between the generated image and the input caption, such as the CLIPScore~\cite{hessel2021clipscore}. However, as CLIP struggles with compositional texts~\cite{kamath2023text}, recent studies such as VPEval~\cite{vpeval} and LLMScore~\cite{llmscore} combine VLMs with LLMs like ChatGPT to more accurately score image-text alignment. Most recently, TIFA~\cite{tifa}, VQ2~\cite{vq2}, and Davidsonian~\cite{davidsonian} use LLMs to generate a set of Q\&A from input captions, then score the image based on the accuracy of a VQA model. \autoref{app:other} describes these methods in details. \autoref{tab:llava} shows that VisualGPTScore (and its debiased $\alpha$=1 version) outperforms such complex approaches for image-text alignment, needing only an OTS state-of-the-art VLM, LLaVA-1.5~\cite{llava}. This suggests that image-conditioned language models can already serve as robust alignment metrics. We also encourage readers to explore our latest research on VQAScore~\cite{lin2024evaluating, li2024evaluating}, which adapts VisualGPTScore to more advanced generative models trained with visual-question-answering (VQA) datasets.

\section{Discussion and Limitations}
{\bf Summary.} Our study shows the efficacy of {\em generative} pre-training scores in solving {\em discriminative} tasks. We present a first-principles analysis to account for mismatching distributions over text between train and test data. Our analysis motivates a training-free (zero-shot) solution to effectively debias language priors in generative scores. We hope our analysis can encourage future work to revisit the issue of language biases in vision-language benchmarks.

{\bf Limitations and future work.} VisualGPTScore depends on VLMs pre-trained on noisy and imbalanced web data, which may result in biases~\cite{mehrabi2021survey, parashar2024neglected}. We make several simplified assumptions in the main paper to offer an intuitive explanation of VisualGPTScore. For instance, the image-conditioned language model might not accurately represent $P_{train}(\mathbf{t}|\mathbf{i})$ and assigns higher scores towards more common texts. We examine this phenomenon in \autoref{app:biased}. Future work may attempt other sampling methods like coreset selection~\cite{guo2022deepcore, wu2023multimodal} to estimate $P_{train}(\mathbf{t})$ with improved efficiency. As VisualGPTScore shows competitive performance, distilling it into discriminative CLIPScore~\cite{miech2021thinking} can reduce its inference cost. Finally, VQAScore~\cite{lin2024evaluating, li2024evaluating} applies VisualGPTScore to the latest vision-language models trained on visual-question-answering (VQA) datasets to achieve the state-of-the-art performance. This demonstrates that generative scoring is a more reliable alternative to CLIPScore~\cite{hessel2021clipscore} for automated evaluation of text-to-image models.

\section*{Impact Statement}

VisualGPTScore is developed with the important goal of advancing the field of vision-language models. It has many positive societal impacts, such as improving the scientific evaluation of generative models~\cite{lin2024evaluating, li2024evaluating}. Nonetheless, we encourage future work to study its biases, especially since the underlying models are trained on noisy and imbalanced data~\cite{parashar2024neglected, mehrabi2021survey}.

\clearpage
\bibliography{example_paper}
\bibliographystyle{icml2024}

\newpage
\appendix
\onecolumn
\section{Is VisualGPTScore a Biased Estimator of $P_{train}(\mathbf{t}|\mathbf{i})$?}
\label{app:biased}
{\bf Retrieval performance on trainset (LAION).} This paper is built on the assumption that VisualGPTScore is a reliable estimator of $P_{train}(\mathbf{t}|\mathbf{i})$. However, this simplifying assumption does not completely hold for the BLIP model we examine. We speculate that such OTS generative scores are biased towards more common texts. We witness this same phenomenon in \autoref{tab:laion_retrieval_all}, where we perform image-text retrieval on random subsets from training distribution LAION-114M~\cite{blip}.

\begin{table}[h]
  \centering
  \caption{ {\bf Retrieval performance on randomly sampled training (LAION114M) subsets with varied sizes.} Table (a) shows that while OTS generative scores are robust for T-to-I retrieval, its performance degrades on I-to-T retrieval tasks when the number of candidate texts increases. This implies that OTS generative scores suffer from language biases towards certain texts even in the training set. Nonetheless, we show that our debiasing solution using either $\alpha = 1$ or optimal $\alpha^{\ast} \in [0, 1]$ with a step size of 0.001, can consistently boost the performance. Figure (b) visualizes $\alpha$-debiasing results on LAION subsets, where each curve represents a different sample size.}
  \scalebox{0.67}{
    \begin{tabular}[h]{cc}
        \begin{NiceTabular}{llllllll}
            \CodeBefore
              \rectanglecolor{softyellow}{4-3}{9-3}
              \rectanglecolor{softpink}{4-4}{9-4}
              \rectanglecolor{softgreen}{4-5}{9-5}
            \Body
            \toprule
            \multirow{3}{*}{Dataset Size} & \multicolumn{5}{c}{I-to-T Retrieval} & \multicolumn{2}{c}{T-to-I Retrieval}
            \\ \cmidrule(lr){2-6} \cmidrule(lr){7-8}
            & \multirow{2}{*}{ITM} & \multicolumn{4}{c}{$\frac{P_{train}(\mathbf{t}|\mathbf{i})}{P_{train}(\mathbf{t})^{\alpha}}$} & \multirow{2}{*}{ITM} & \multirow{2}{*}{$P_{train}(\mathbf{t}|\mathbf{i})$} \\ \cmidrule{3-6} 
            & & $\alpha$=0 & $\alpha$=1 & $\alpha$=$\alpha^{\ast}$ & $\alpha^{\ast}$  \\
            \midrule
            100   & \textbf{96.0} & 59.0 & 94.0 & \textbf{95.0} & 0.535 & 95.0 & \textbf{97.0} \\
            1000   & \textbf{90.9} & 37.1 & 71.7 & 85.7 & 0.733 & 92.0 & \textbf{93.1} \\
            2000   & \textbf{87.2} & 32.8 & 62.3 & 64.3 & 0.840 & 87.8 & \textbf{89.8}\\
            5000   & \textbf{79.8} & 25.1 & 50.9 & 54.1 & 0.727 & 81.9 & \textbf{84.4}\\
            \bottomrule
        \end{NiceTabular}
        &
        \begin{tabular}[h]{l}
        \includegraphics[width=0.45\textwidth]{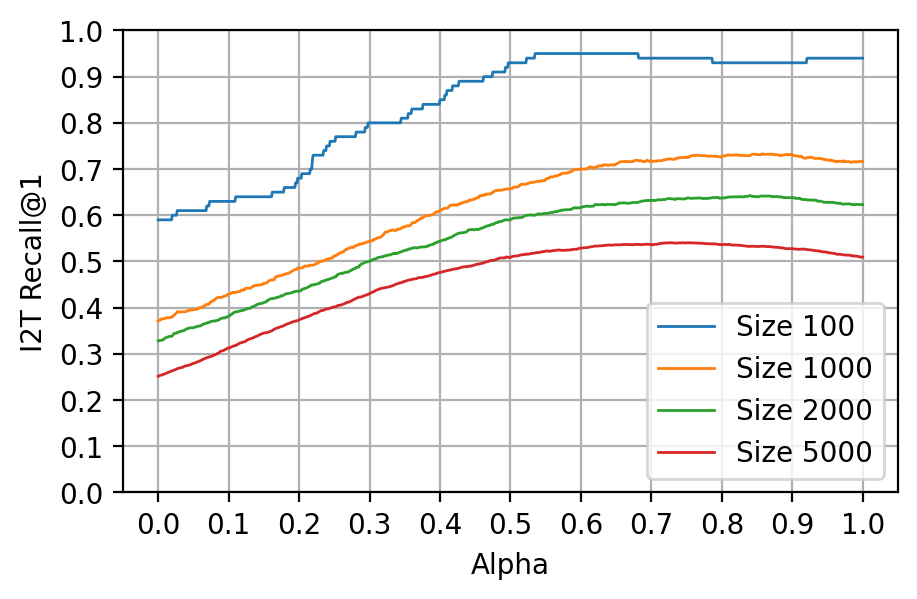}
        \end{tabular} \\
        \multicolumn{1}{c}{{\large (a) Performance on LAION trainset retrieval}} & \multicolumn{1}{c}{{\large (b) Alpha-tuning on LAION}}
    \end{tabular}
    }
    
    \label{tab:laion_retrieval_all}
\end{table}

{\bf Modelling the language bias in VisualGPTScore.} As evidenced in \autoref{tab:laion_retrieval_all}, we believe VisualGPTScore is biased towards more common texts due to modelling error. To consider this error in our analysis, we rewrite the VisualGPTScore as:
\begin{equation}
    \textbf{VisualGPTScore}(\mathbf{t}, \mathbf{i}) := \hat{P}_{train}(\mathbf{t} | \mathbf{i}) = P_{train}(\mathbf{t} | \mathbf{i}) \cdot P_{train}(t)^{\beta},
\end{equation}
where $\hat{P}$ represents the (biased) model estimate and $P$ represents the true distribution. The model bias towards common texts is encoded by an unknown parameter $\beta$.

{\bf Monte Carlo estimation using $\hat{P}$.} Because our Monte Carlo sampling method relies on $\hat{P}_{train}(\mathbf{t}|\mathbf{i})$, it is also a biased estimator of $P_{train}(\mathbf{t})$:

\begin{equation}
    \hat{P}_{train}(\mathbf{t}) := \frac{1}{n} \sum_{k=1}^n \hat{P}_{train}(\mathbf{t} | \mathbf{i}_k) = P_{train}(\mathbf{t})^{1+\beta}.
\end{equation}

{\bf Rewriting optimal I-to-T objective with $\hat{P}$.} We can rewrite \autoref{eq:optimal_i2t} as:
\begin{align} 
    P_{test}(\mathbf{t}|\mathbf{i}) &\propto {P}_{train}(\mathbf{t}|\mathbf{i}) \frac{P_{test}(\mathbf{t)}}{P_{train}(\mathbf{t})} \\
    &= \hat{P}_{train}(\mathbf{t}|\mathbf{i}) \frac{P_{test}(\mathbf{t)}}{P_{train}(\mathbf{t})^{1+\beta}} \\
    &= \hat{P}_{train}(\mathbf{t} | \mathbf{i}) \frac{P_{test}(\mathbf{t)}}{\hat{P}_{train}(\mathbf{t})} 
\label{eq:optimal_i2t_biased_version}
\end{align}

{\bf $\alpha$-debiasing with $\hat{P}$.} Using \autoref{eq:optimal_i2t_biased_version}, we can reformulate $\alpha$-debiasing (\autoref{eq:debias_i2t}) as follows:

\begin{align}
P_{test}(\mathbf{t}) \propto P_{train}(\mathbf{t})^{1-\hat{\alpha}} \quad \Rightarrow \text{Optimal score is }& \frac{\hat{P}_{train}(\mathbf{t} | \mathbf{i})}{\hat{P}_{train}({\bf t})^{\alpha}}
\label{eq:debias_i2t_bias}
\end{align}
where $\alpha = \frac{\hat{\alpha} + \beta}{1 + \beta}$. Notably, the above equation has the same structure as before (\autoref{eq:debias_i2t}). This implies that even if $P_{train}(\mathbf{t}) = P_{test}(\mathbf{t})$, we still anticipate $\alpha = \frac{\beta}{1+\beta} \neq 0$. This accounts for why the optimal $\alpha$ is not 0 when we perform I-to-T retrieval on trainset in \autoref{tab:laion_retrieval_all}. 

{\bf Implication for vision-language modelling.} Our analysis indicates that similar to generative LLMs~\cite{pmimt1, pmimt2}, contemporary image-conditioned language models also experience issues related to imbalanced learning~\cite{kang2019decoupling}. Potential solutions could be: (a) refined sampling techniques for Monte Carlo estimation of $P(\mathbf{t})$ such as through dataset distillation~\cite{wu2023multimodal}, and (b) less biased modelling of $P(\mathbf{t}|\mathbf{i})$ such as through controllable generation~\cite{keskar2019ctrl}.

\section{Ablation Studies on $\alpha$-Debiasing}
\label{app:alpha_ablation}
{\bf Details of Gaussian noise samples. } BLIP and BLIP-2 experiments sample Gaussian noise images with a mean of $1.0$ and a standard deviation of $0.25$. By default, we use 100 images for Winoground, 30 images for EqBen, 1 image for ImageNet, and 3 images for the rest of the benchmarks. 

{\bf Estimating $P_{train}(\textbf{t})$ via Gaussian noise images is more sample-efficient.} We use Winoground to show that sampling Gaussian noise images to calculate $P_{train}(\mathbf{t})$ can be more efficient than sampling trainset images. As demonstrated in \autoref{tab:sampling_test}, a limited number of Gaussian noise images (e.g., 3 or 10) can surpass the results obtained with 1000 LAION images. Moreover, using null images produces less variance in the results.

\begin{table}[h]
\caption{{\bf Comparing sampling of Gaussian noise images and trainset images 
for estimating $P_{train}(\mathbf{t})$.} We report text scores of $\alpha$-debiasing on Winoground I-to-T retrieval task. We ablate 3/10/100/1000 Gaussian noise and LAION samples and report both mean and std using 5 sampling seeds. The optimal $\alpha^{\ast} \in [0, 1]$ is searched on testset via a step size of $0.001$. The Gaussian noise images are sampled with a mean calculated from the LAION subset and a fixed std of $0.25$. }
\centering
\scalebox{0.9}{
\begin{tabular}{lcc|cc}
\toprule
\multirow{2}{*}{Sample Size} & \multicolumn{2}{c}{Guassian Noise Images} & \multicolumn{2}{c}{Trainset Images} \\
\cmidrule(lr){2-3} \cmidrule(lr){4-5} & $\alpha$=$\alpha^{\ast}_{test}$ & $\alpha^{\ast}_{test}$ & $\alpha$=$\alpha^{\ast}_{test}$ & $\alpha^{\ast}_{test}$ \\
\midrule
3 & $35.95_{(0.5)}$& 0.821$_{(0.012)}$ &  $32.20_{(1.6)}$& 0.706$_{(0.150)}$ \\
10 & $36.25_{(0.4)}$& 0.827$_{(0.016)}$ & $33.60_{(0.9)}$& 0.910$_{(0.104)}$ \\
100 & $36.35_{(0.1)}$& 0.840$_{(0.010)}$ & $34.70_{(0.6)}$& 0.910$_{(0.039)}$ \\
1000 & $36.25_{(0.0)}$& 0.850$_{(0.000)}$ & $35.15_{(0.3)}$& 0.960$_{(0.033)}$\\
\bottomrule

\end{tabular}
}
\label{tab:sampling_test}

\end{table}

{\bf Alternative approach on COCO/Flickr30k: estimating $P_{train}(\mathbf{t})$ using testset images. } For large-scale retrieval benchmarks like COCO~\cite{coco} and Flickr30k~\cite{flickr30k}, we can directly average scores of all candidate images (in the order of thousands) to efficiently approximate $P_{train}(\mathbf{t})$ without the need to sample any Gaussian noise images. This approach incurs zero computation cost as we have already pre-computed scores between each candidate image and text. We show in \autoref{tab:retrieval_Gaussian_vs_testset} that using testset images indeed results in better performance than sampling 3 Gaussian noise images.

\begin{table}[h]
  \centering
  \caption{ {\bf I-to-T retrieval on COCO/Flickr30k using different sampling methods.} Estimating $P_{train}(\mathbf{t})$ by averaging the scores of testset images (with zero computational cost) demonstrates superior performance compared to sampling additional Gaussian noise images.}
  \scalebox{0.79}{
    \begin{tabular}[h]{cc}
        \multicolumn{2}{c}{
            \begin{tabular}[h]{llccccc}
                \toprule
                \multirow{2}{*}{Metric} & \multirow{2}{*}{Benchmark}
                & \multirow{2}{*}{$P_{train}(\mathbf{t}|\mathbf{i})$} & \multirow{2}{*}{Sampling Method} & \multicolumn{3}{c}{$\frac{P_{train}(\mathbf{t}|\mathbf{i})}{P_{train}(\mathbf{t})^{\alpha}}$} \\ \cmidrule{5-7} 
                & & & & $\alpha$=1 & $\alpha$=$\alpha^{\ast}_{val}$ & $\alpha^{\ast}_{val}$  \\
                \midrule
                \multirow{4}{*}{{\small R@1 / R@5}}& \multirow{2}{*}{COCO}   & \multirow{2}{*}{19.7 / 40.6} & Testset Images & 46.2 / 73.1 & 48.0 / 74.2 & 0.819 \\
                & & & Null Images  & 24.4 / 52.6 & 40.4 / 66.6 & 0.600 \\ \cmidrule{2-7} %
                & \multirow{2}{*}{Flickr30k}   & \multirow{2}{*}{34.6 / 59.0} & Testset Images & 58.7 / 88.0 & 63.6 / 89.2 & 0.719 \\
                & & & Null Images & 27.8 / 62.2 & 48.5 / 79.0 & 0.427 \\ %

                \bottomrule
            \end{tabular} 
        }
    \end{tabular}
    }
    \label{tab:retrieval_Gaussian_vs_testset}
\end{table}

{\bf Tuning $\alpha$ with a valset.} In \autoref{tab:coco_flickr_curve}, similar performance trends are observed across validation and test splits of COCO and Flickr30k I-to-T retrieval benchmarks using the same $\alpha \in [0, 1]$. Furthermore, $\alpha_{test}^{\ast}$ and $\alpha_{val}^{\ast}$ are empirically close. As such, our method can function as a reliable training-free debiasing method.

\begin{table}[h]
    \centering
    \caption{ {\bf $\alpha$-debiasing results on both the valset and testset for COCO/Flickr30k I-to-T retrieval.} We observe that validation and test performance are strongly correlated while we interpolate $\alpha \in [0, 1]$.}
    \scalebox{0.68}{
        \begin{tabular}{cc}
                \includegraphics[width=0.49\textwidth]{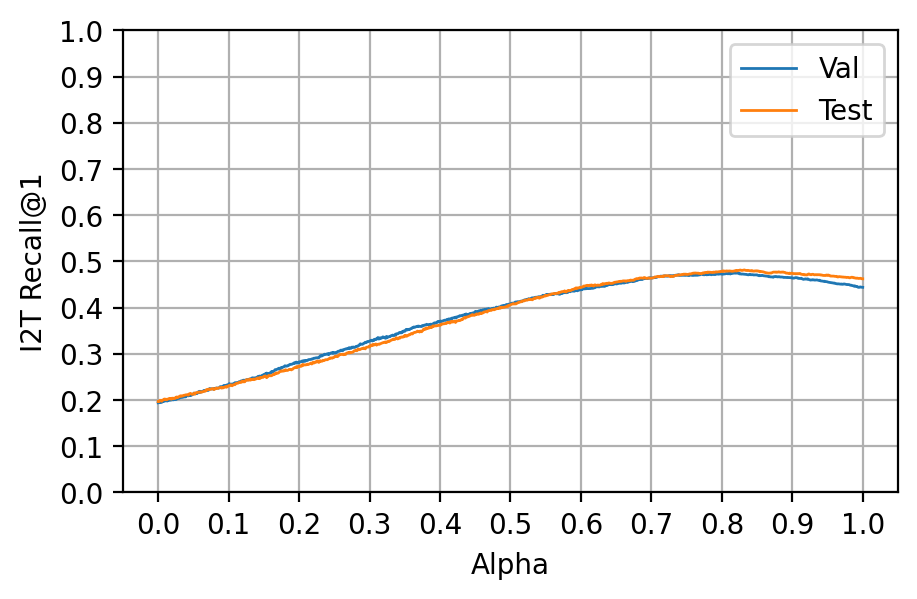} &
                \includegraphics[width=0.49\textwidth]{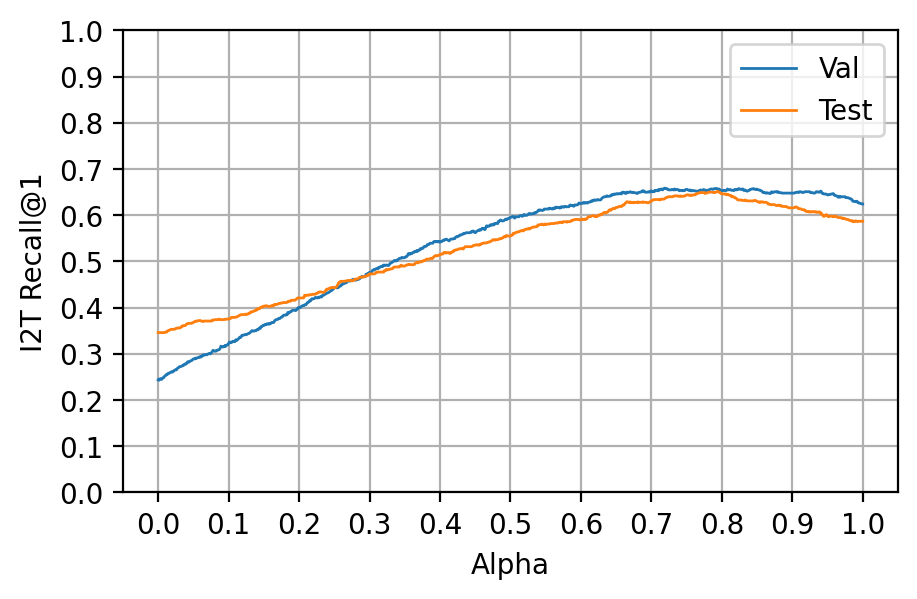}
            \\
            \multicolumn{1}{c}{(b) Alpha-tuning on COCO Retrieval} & \multicolumn{1}{c}{(c) Alpha-tuning on Flickr Retrieval}
        \end{tabular}
    }
    
    \label{tab:coco_flickr_curve}
\end{table}

\section{Experiments with BLIP-2}
\label{app:blip_2}
We provide BLIP-2 results for completeness.

{\bf BLIP-2~\cite{blip2} overview.} BLIP-2 leverages frozen pre-trained image encoders~\cite{fang2022eva} and large language models~\cite{flan, opt} to bootstrap vision-language pre-training. It proposes a lightweight Querying Transformer (Q-Former) that is trained in two stages. Similar to BLIP~\cite{blip}, Q-Former is a mixture-of-expert model that can calculate ITC, ITM, and captioning loss given an image-text pair. Additionally, it introduces a set of trainable query tokens, whose outputs serve as {\em visual soft prompts} prepended as inputs to LLMs. In its first training stage, Q-Former is fine-tuned on the same LAION dataset using the same objectives (ITC+ITM+captioning) as BLIP. In the second stage, the output query tokens from Q-Former are fed into a frozen language model, such as FLAN-T5~\cite{flan} or OPT~\cite{flan}, after a linear projection trained only with captioning loss. BLIP-2 achieves state-of-the-art performance on various vision-language tasks with significantly fewer trainable parameters.

{\bf BLIP-2 results (\autoref{tab:blip2_vanilla} and \autoref{tab:blip2_other_benchmarks}).} We present retrieval performance of the BLIP-2 model that uses ViT-L as the frozen image encoder. We report results for both the first-stage model (denoted as Q-Former) and the second-stage model which employs FLAN-T5~\cite{flan} as the frozen LLM. Our $\alpha$-debiasing solutions generalize to all variants of BLIP-2. %

\begin{table}[h]
\centering
\caption{\small {\bf BLIP-2 on ARO/Crepe/VL-CheckList/SugarCrepe.}}
\scalebox{0.7}{
\begin{tabular}{llccccc}
\toprule
\multirow{2}{*}{Benchmark} & \multirow{2}{*}{Dataset} & \multirow{2}{*}{Random}  & \multicolumn{3}{c}{w. Q-Former} & w. Flan-T5 \\ 
\cmidrule(lr){4-6} \cmidrule(lr){7-7} & & & ITC & ITM & $P_{train}(\mathbf{t}|\mathbf{i})$ & $P_{train}(\mathbf{t}|\mathbf{i})$ \\
\midrule
\multirow{4}{*}{ARO} & VG-Relation     & 50.0 & 46.4 & 67.2 & 90.7 & 89.1 \\
& VG-Attribution  & 50.0 & 76.0 & 88.1 & 94.3  & 90.9 \\
& COCO-Order      & 20.0 & 28.5 & 25.2 & 96.8 & 99.3 \\
& Flickr30K-Order & 20.0 & 25.3 & 28.6 & 97.5 & 99.7 \\ 
\midrule
\multirow{3}{*}{Crepe} & Atom-Foils      & 16.7 & 20.8 & 20.9 & 74.7 & 69.7 \\
& Negate          & 16.7 & 13.4 & 14.2 & 79.1 & 90.0 \\
& Swap            & 16.7 & 13.4 & 18.0 & 79.5 & 79.1\\ 
\midrule
VL-CheckList & Object   & 50.0 & 89.7 & 89.2 & 90.1 & 84.1 \\ 
VL-CheckList & Attribute   & 50.0 & 76.6 & 79.3 & 73.9 & 70.6 \\ 
VL-CheckList & Relation   & 50.0 & 70.5 & 72.3 & 89.9 & 56.7 \\ 
\midrule
SugarCrepe & Replace   & 50.0 & 86.7 & 88.5 & 93.0 & 82.4 \\ 
SugarCrepe & Swap   & 50.0 & 69.8 & 80.9 & 91.2 & 80.8 \\ 
SugarCrepe & Add   & 50.0 & 86.5 & 88.0 & 92.7 & 76.2 \\ 
\bottomrule
\end{tabular}
}
\label{tab:blip2_vanilla}
\end{table}

\begin{table}[h]
\centering
\caption{\small {\bf BLIP-2 on Winoground/EqBen.} 
}
\scalebox{0.7}{
\begin{NiceTabular}{lllllllllll}
        \CodeBefore
          \rectanglecolor{softyellow}{4-5}{9-5}
          \rectanglecolor{softpink}{4-6}{9-6}
          \rectanglecolor{softgreen}{4-7}{9-7}
        \Body
\toprule
\multirow{3}{*}{Benchmark} & \multirow{3}{*}{Model} & \multicolumn{6}{c}{I-To-T (Text Score)} & \multicolumn{3}{c}{T-To-I (Image Score)} \\ \cmidrule(lr){3-8} \cmidrule(lr){9-11}
&  & \multirow{2}{*}{ITC} & \multirow{2}{*}{ITM} & \multicolumn{4}{c}{$\frac{P_{train}(\mathbf{t}|\mathbf{i})}{P_{train}(\mathbf{t})^{\alpha}}$} & \multirow{2}{*}{ITC} & \multirow{2}{*}{ITM} & \multirow{2}{*}{$P_{train}(\mathbf{t}|\mathbf{i})$} \\
\cmidrule(lr){5-8} 
 & & & & $\alpha$=0 & $\alpha$=1 & $\alpha$=$\alpha^{\ast}$ & $\alpha^{\ast}$ &  & \\
\midrule
\multirow{3}{*}{Winoground} & BLIP & 28.0 & 35.8 & 27.0 & 33.0 & 36.5 & 0.836 & 9.0 & 15.8 & 21.5 \\
& BLIP2-QFormer & 30.0 & 42.5 & 24.3 & 29.3 & 33.0& 0.882 & 10.5 & 19.0 & 20.0\\
& BLIP2-FlanT5 & - & - & 25.3 & 31.5 & 34.3 & 0.764 & - & - & 19.5 \\
\midrule
\multirow{3}{*}{EqBen (Val)} & BLIP & 20.9 & 26.0 & 9.6 & 19.8 & 19.8 & 0.982 & 20.3 & 20.3 & 26.1 \\
& BLIP2-QFormer & 32.1 & 36.2 & 12.2 & 21.9 & 22.2 & 0.969 & 23.4 & 28.4 & 26.6 \\
& BLIP2-FlanT5 & - & - & 8.5 & 22.0 & 22.0 & 1.000 & - & - & 20.9 \\
\bottomrule
\end{NiceTabular}
}
\label{tab:blip2_other_benchmarks}
\end{table}

\section{Additional Reports}
\label{app:other}

{\bf Computational resources.} All experiments use a single NVIDIA GeForce 3090s GPU. 

{\bf Details of \autoref{tab:i_to_t_retrieval}.} For CLIP~\cite{clip}, LAION2B-CLIP, and LAION5B-CLIP~\cite{laion5b}, we report the results from \citet{sugarcrepe} using the ViT-B-32, ViT-bigG-14, and xlm-roberta-large-ViT-H-14 models respectively. The results of NegCLIP~\cite{aro}, Structure-CLIP~\cite{huang2023structure}, SVLC~\cite{doveh2022teaching}, SGVL~\cite{herzig2023incorporating}, DAC-LLM, and DAC-SAM~\cite{doveh2023dense} are directly copied from their original papers. We run BLIP-ITC and BLIP-ITM using our own codebase, which will be released to the public.

{\bf Method descriptions for \autoref{tab:llava}.} CLIPScore~\cite{hessel2021clipscore} measures the cosine similarity (dot product) score between an image and text, each embedded using the CLIP image and text encoder, respectively. VPEval~\cite{vpeval} utilizes GPT-3.5 to translate the text prompt into a Python-like program that invokes vision foundation models such as CLIP, BLIP, and GroundingDINO, to examine fine-grained image details. LLMScore~\cite{llmscore} uses BLIP-2 to first caption the image, then uses ChatGPT to score the difference between the BLIP-generated caption and the text prompt. TIFA~\cite{tifa} and Davidsonian~\cite{davidsonian} first use LLMs such as a finetuned Llama-2 or GPT-3.5 to generate a set of Q\&A given the text prompt, then return the accuracy score of the VQA model. VQ2~\cite{vq2} uses a finetuned FlanT5 to generate the Q\&A, then averages the log likelihoods of the generated answers.

{\bf Implementation details of \autoref{tab:llava}.} We report the performance on Winoground~\cite{winoground} and EqBen-Mini, which is an official subset of EqBen~\cite{eqben} for benchmarking large foundational VLMs. We follow the official implementation of CLIPScore~\cite{hessel2021clipscore} to report the performance of CLIP-ViT-B-32~\cite{clip}. For VPEval~\cite{vpeval} and LLMScore~\cite{llmscore}, we strictly follow the official codebase to benchmark their performance. For TIFA~\cite{tifa}, VQ2~\cite{vq2}, Davidsonian~\cite{davidsonian}, we strictly follow their released code and adopt their QA-generation language models (or in-context Q\&A samples for ChatGPT). However, as we do not have access to the private VQA models they adopted, e.g., PaLI-17B, we implement these approaches using LLaVA-1.5-13B~\cite{llava} as the VQA model. We stick to the default system message to prompt LLaVA-1.5, which can be found on their official GitHub repo. For fair comparison, our VisualGPTScore is also implemented using LLaVA-1.5-13B. We only use the system message without appending any questions when computing $P(\text{text}|\text{image})$. For $\alpha$-debiasing, we sample a single Gaussian image with a mean of 0 and standard deviation of 0.25 (derived from the statistics of training images used to train LLaVA).

{\bf Group scores on Winoground/EqBen using BLIP (\autoref{tab:blip_groupscore}).} 

\begin{table}[h]
\centering
\caption{\small Performance comparison of BLIP's ITCScore, ITMScore, and $\alpha$-tuned VisualGPTScore$^{\alpha^{\ast}}$ on Winoground and EqBen. }
\scalebox{0.8}{

\begin{tabular}{lccc|ccc}
\toprule
\multirow{2}{*}{Method} & \multicolumn{3}{c}{Winoground (all)} & \multicolumn{3}{c}{EqBen (val)} \\
\cmidrule(lr){2-4} \cmidrule(lr){5-7} & Text Score & Image Score & Group Score & Text Score & Image Score & Group Score \\
\midrule
ITCScore & 28.0 & 9.0 & 6.5 &  20.9 & 20.3 & 10.6 \\
ITMScore & 35.8 & 15.8 & 13.3 &  26.0 &20.3 &12.6 \\
VisualGPTScore$^{\alpha^{\ast}}$ & 36.5 &21.5 & 16.8 &20.4 &26.1 &11.7 \\
\bottomrule
\end{tabular}
}
\label{tab:blip_groupscore}
\end{table}

{\bf Fine-grained tags on Winoground (\autoref{tab:finegrained_winoground_blip}).} 

\begin{table}[h]
\centering
\caption{\small BLIP performance on Winoground subtags~\cite{diwan2022winoground}. We report the number of test instances for each subtag and their respective text score, image score, group score.}
\scalebox{0.9}{
\begin{tabular}{lllccc}
\toprule
Dataset & Size & Method & Text Score & Image Score & Group Score \\
\midrule
\multirow{3}{*}{NoTag} & \multirow{3}{*}{171} & ITCScore & 32.6 & 11.6 & 8.1 \\
& & ITMScore &41.9 & 21.5& 19.2\\
& & VisualGPTScore$^{\alpha^{\ast}}$ & 43.0 & 28.5 & 23.8\\
\midrule

\multirow{3}{*}{NonCompositional} & \multirow{3}{*}{30} & ITCScore & 43.3 & 16.7 & 16.7\\
& & ITMScore & 50.0& 23.3& 16.7\\
& & VisualGPTScore$^{\alpha^{\ast}}$ & 43.3& 33.3 & 26.7 \\
\midrule

\multirow{3}{*}{AmbiguouslyCorrect} & \multirow{3}{*}{46} & ITCScore & 32.6 & 8.7 &6.5 \\
& & ITMScore &28.3 & 6.5& 2.2\\
& & VisualGPTScore$^{\alpha^{\ast}}$ &26.1 & 19.6& 8.7\\
\midrule

\multirow{3}{*}{VisuallyDifficult} & \multirow{3}{*}{38} & ITCScore & 29.0& 7.9& 7.9\\
& & ITMScore & 26.3 & 10.5& 7.9\\
& & VisualGPTScore$^{\alpha^{\ast}}$ & 31.6& 13.2& 7.9\\
\midrule

\multirow{3}{*}{UnusualImage} & \multirow{3}{*}{56} & ITCScore &32.5 &8.9 & 8.9\\
& & ITMScore & 21.4 &10.7 & 7.1 \\
& & VisualGPTScore$^{\alpha^{\ast}}$ &30.4 & 10.7 & 8.9\\
\midrule

\multirow{3}{*}{UnusualText} & \multirow{3}{*}{50} & ITCScore & 20.0 & 8.0 & 6.0\\
& & ITMScore & 38.0 & 12.0& 12.0\\
& & VisualGPTScore$^{\alpha^{\ast}}$ &30.0 & 18.0& 12.0\\
\midrule

\multirow{3}{*}{ComplexReasoning} & \multirow{3}{*}{78} & ITCScore & 16.7 & 2.6 & 1.3 \\
& & ITMScore & 21.8 & 5.1 & 2.6\\
& & VisualGPTScore$^{\alpha^{\ast}}$ & 21.8& 10.3& 6.4\\

\bottomrule

\end{tabular}
}
\label{tab:finegrained_winoground_blip}
\end{table}
{\bf Performance on SugarCrepe (\autoref{tab:sugar_crepe}).} 

\begin{table}[h!]
\centering
\caption{\small {\bf Performance on SugarCrepe~\cite{sugarcrepe}.} SugarCrepe is the most recent visio-linguistic compositionality benchmark which improves upon previous Crepe~\cite{crepe} by using state-of-the-art large language models (including ChatGPT), instead of rule-based templates, to generate more natural negative text captions. We show that text-only baselines and LLM-based methods indeed fail to succeed on SugarCrepe. However, our OTS generative approaches still achieve competitive results compared against SOTA discriminative approaches. The results of human performance, text-only baseline, and SOTA CLIP and NegCLIP-SugarCrepe are directly taken from the ~\citet{sugarcrepe}. For other approaches, we evaluate their performance following the same procedure as described in main texts.}
\scalebox{0.9}{
\begin{NiceTabular}{llccc|c}
\CodeBefore
          \rectanglecolor{softred}{7-3}{9-6}
          \rectanglecolor{softblue}{10-3}{10-6}
          \rectanglecolor{softyellow}{18-3}{20-6}
          \rectanglecolor{softgreen}{21-3}{23-6}
        \Body
\toprule

\multirow{2}{*}{Method} & \multirow{2}{*}{Model} & \multicolumn{4}{c}{SugarCrepe} \\
\cmidrule(lr){3-6}
& & {\bf Replace} & {\bf Swap} & {\bf Add} & {\bf AVG} \\
\midrule
Human Performance & - & 98.67 & 99.50 & 99.00 & 99.06 \\
Random Chance & - & 50.00 & 50.00 & 50.00 & 50.00 \\
\hline
\multirow{2}{*}{Text-Only Baseline} & Vera & 49.46 & 49.30 & 49.50 & 49.42 \\
& Grammar & 50.00 & 50.00 & 50.00 & 50.00  \\
\hline
\multirow{3}{*}{$P_{LLM}(\mathbf{t})$} & Bart & 48.41 & 51.93 & 61.16 & 53.83 \\
& Flan-T5 & 51.41 & 57.59 & 40.94 & 49.98 \\
& OPT & 58.53 & 66.58 & 45.78 & 56.96 \\
\hline
\multirow{1}{*}{$P_{train}(\mathbf{t})$} & BLIP &  {75.90} & {77.14} & { 70.89} & {74.64} \\
\hline
\multirow{5}{*}{ITCScore} & CLIP-LAION2B & 86.50 & 68.56 & 88.37 & 81.14 \\
& CLIP-LAION5B & 84.98 & 67.95 & 89.62 & 80.85 \\
& BLIP & 85.76 & 73.79 & 85.66 & 81.74 \\
& BLIP-2 & 86.66 & 69.77 & 86.50 & 80.98 \\
& NegCLIP-SugarCrepe & 88.27 & 74.89 & 90.16 & 84.44 \\
\hline
\multirow{2}{*}{ITMScore} & BLIP & 88.68 & 81.29 & 87.57 & 85.85 \\
 & BLIP2-Qformer & 88.45 & 80.87 & 87.96 & 85.76 \\
\hline
\multirow{3}{*}{$P_{train}(\mathbf{t}|\mathbf{i})$} & BLIP & {\bf 93.33} & {\bf 91.00} & {\bf 90.98} & {\bf 91.77} \\
& BLIP2-Qformer & {\bf 93.00} & {\bf 91.24} & {\bf 92.69} & {\bf 92.31} \\
& BLIP2-FlanT5 & 82.44 & 76.57 & 76.24 & 78.42 \\
\hline
\multirow{3}{*}{$\frac{P_{train}(\mathbf{t}|\mathbf{i})}{P_{train}(\mathbf{t})^{\alpha^\ast}}$} & BLIP &  {\bf 95.09} & {\bf 92.39} & {\bf 97.36} & {\bf 94.95} \\
& BLIP2-Qformer & {\bf 94.62} & {\bf 92.27} & {\bf 97.58} & {\bf 94.82} \\
& BLIP2-FlanT5 & 85.69 & 78.80 & 91.76 & 85.42 \\
\bottomrule
\end{NiceTabular}
}
\label{tab:sugar_crepe}
\end{table}

{\bf $\alpha$-debiasing on ARO/Crepe/SugarCrepe/VL-CheckList (\autoref{tab:datasets_plots}).} 

\begin{table*}[h!]
\centering
\caption{\small {\bf $\alpha$-debiasing results on all I-to-T benchmarks and $P_{train}(\mathbf{t})$ frequency charts.} Increasing $\alpha$ from 0 to 1 hurts performance
on benchmarks with non-sensical negative captions such as ARO and Crepe. These benchmarks can also be largely solved with blind algorithms that avoid looking at images. On the other hand, for benchmarks like SugarCrepe with more balanced $P_{train}(\mathbf{t})$ between positives and negatives, tuning $\alpha$ leads to performance gain.
}
\scalebox{0.8}{
\begin{tabular}{cc|cc}
\toprule
\textbf{Alpha-Tuning} & \textbf{Prior Frequency} & \textbf{Alpha-Tuning} & \textbf{Prior Frequency}  \\
\midrule
\includegraphics[width=0.19\textwidth]{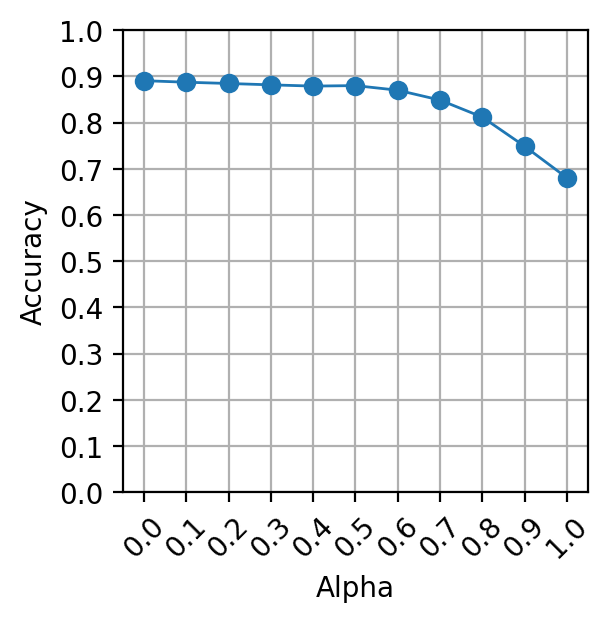} & \includegraphics[width=0.31\textwidth]{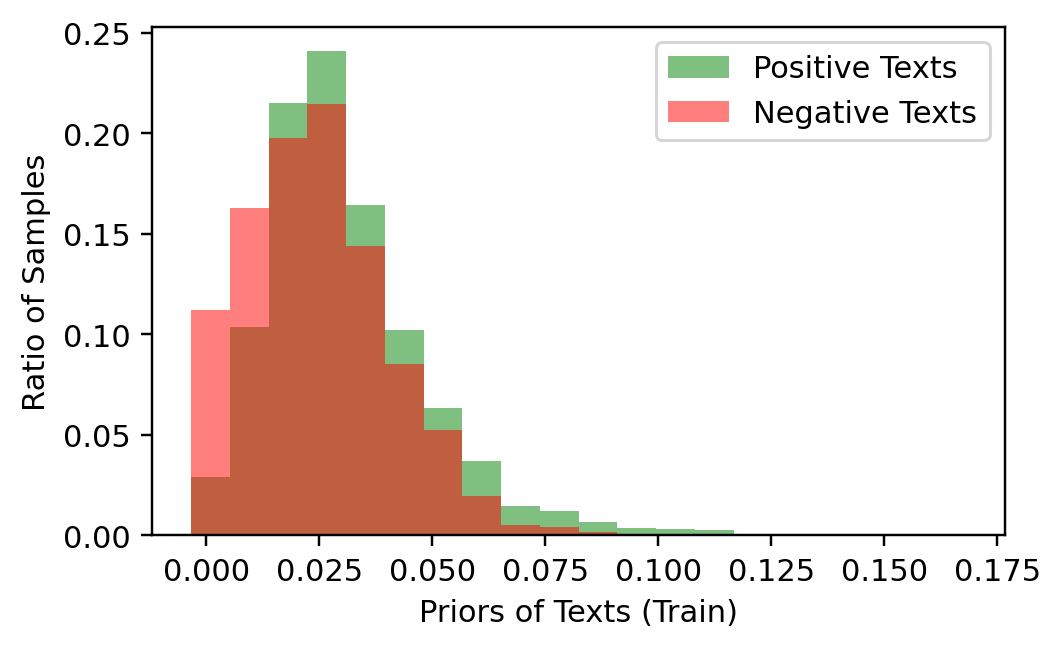} & \includegraphics[width=0.19\textwidth]{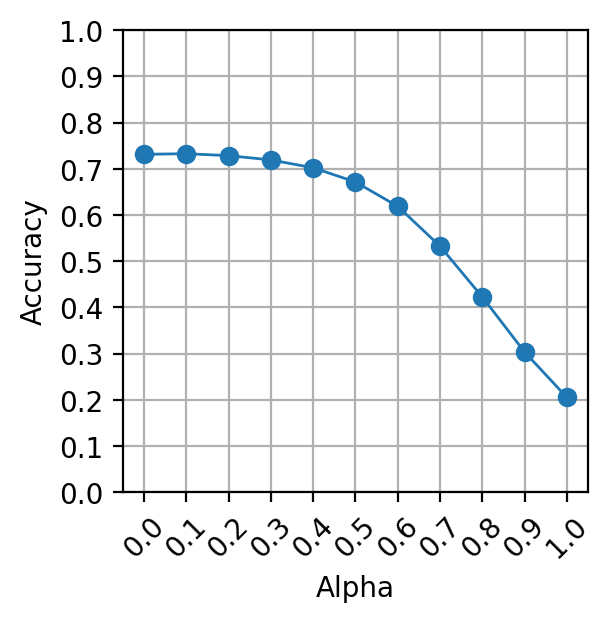} & \includegraphics[width=0.31\textwidth]{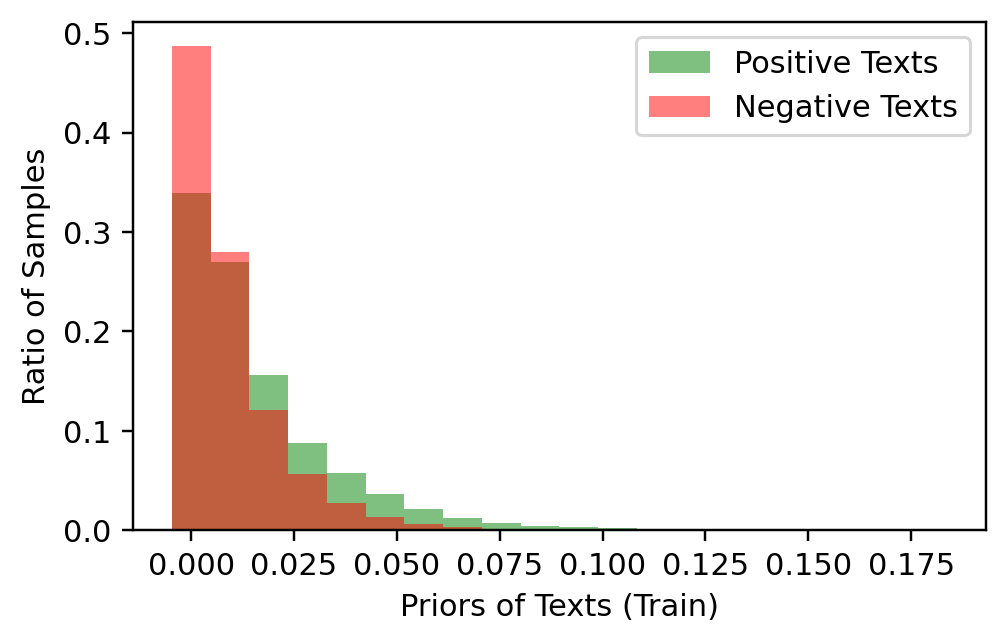} \\
\multicolumn{2}{c|}{ARO (VG-Relation)} & \multicolumn{2}{c}{Crepe (Atom-Foils)} \\
\midrule
\includegraphics[width=0.19\textwidth]{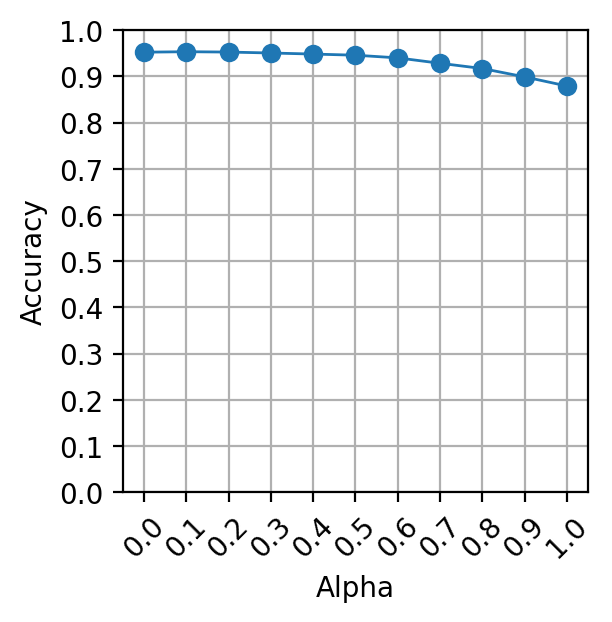} & \includegraphics[width=0.31\textwidth]{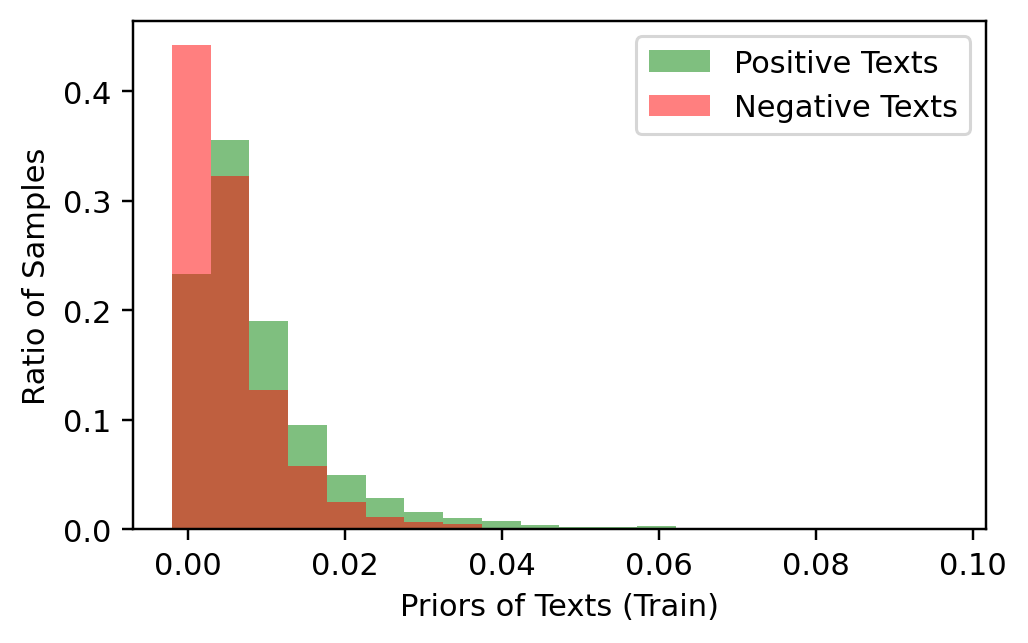} & \includegraphics[width=0.19\textwidth]{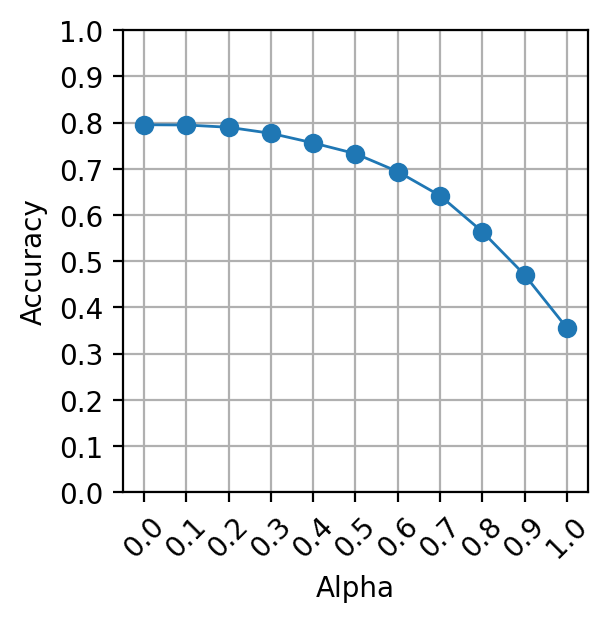} & \includegraphics[width=0.31\textwidth]{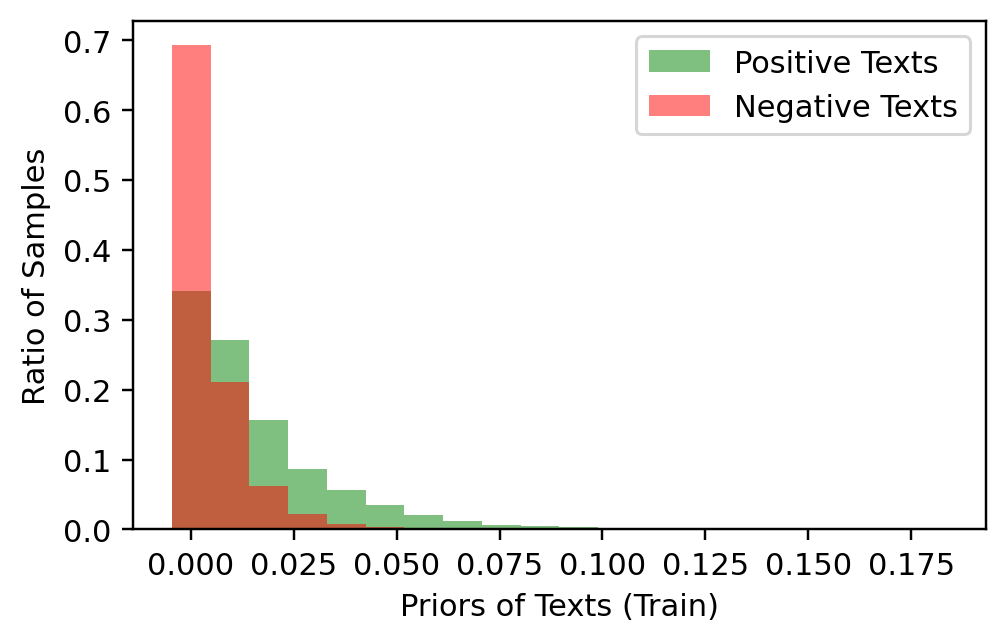} \\
\multicolumn{2}{c|}{ARO (VG-Attribution)} & \multicolumn{2}{c}{Crepe (Negate)} \\
\midrule
\includegraphics[width=0.19\textwidth]{images/diagnosis/flickr_order_alpha.png} & \includegraphics[width=0.31\textwidth]{images/diagnosis/flickr_order_frequency.png} & \includegraphics[width=0.19\textwidth]{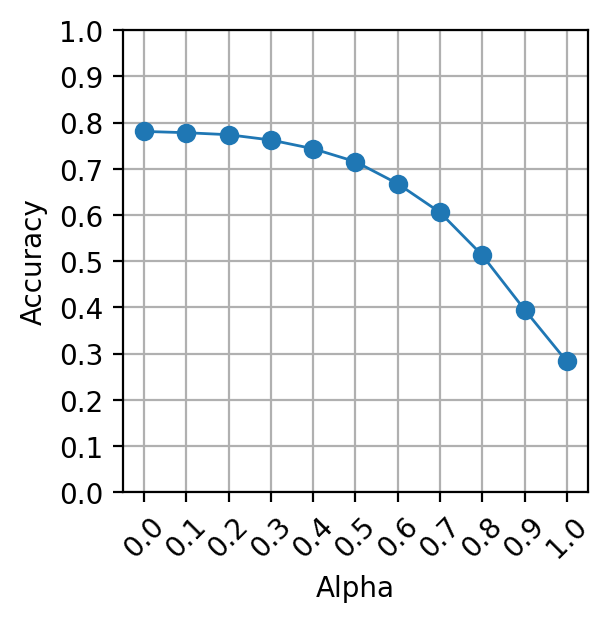} & \includegraphics[width=0.31\textwidth]{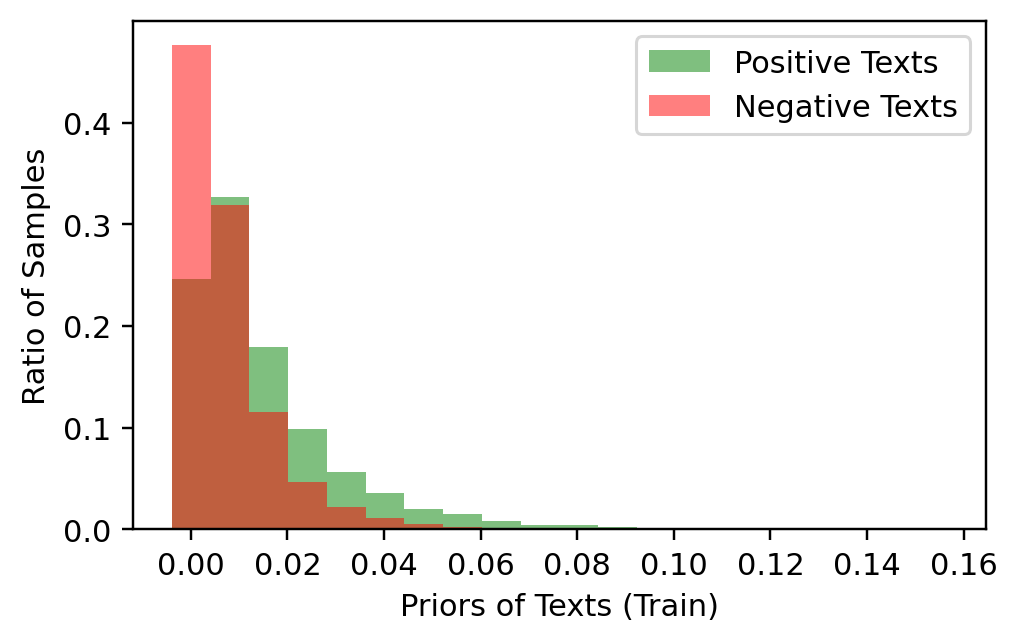} \\
\multicolumn{2}{c|}{ARO (Flickr30K-Order)} & \multicolumn{2}{c}{Crepe (Swap)} \\
\midrule
\includegraphics[width=0.19\textwidth]{images/diagnosis/sugarcrepe_add_alpha.png} & \includegraphics[width=0.31\textwidth]{images/diagnosis/sugarcrepe_add_frequency.png} & \includegraphics[width=0.19\textwidth]{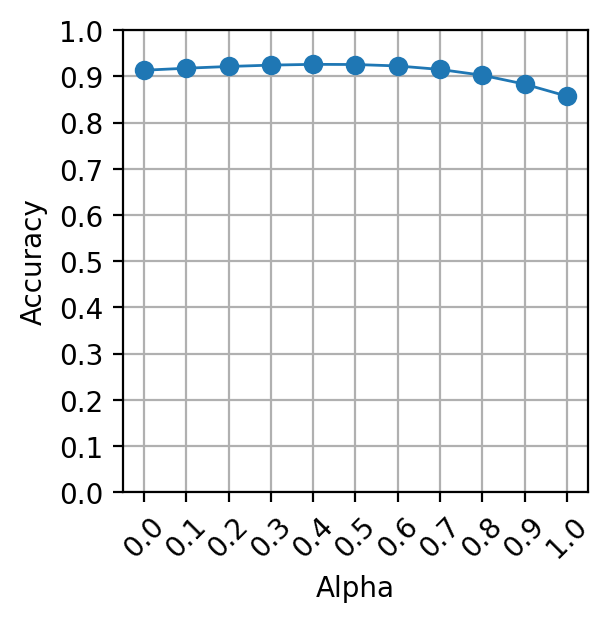} & \includegraphics[width=0.31\textwidth]{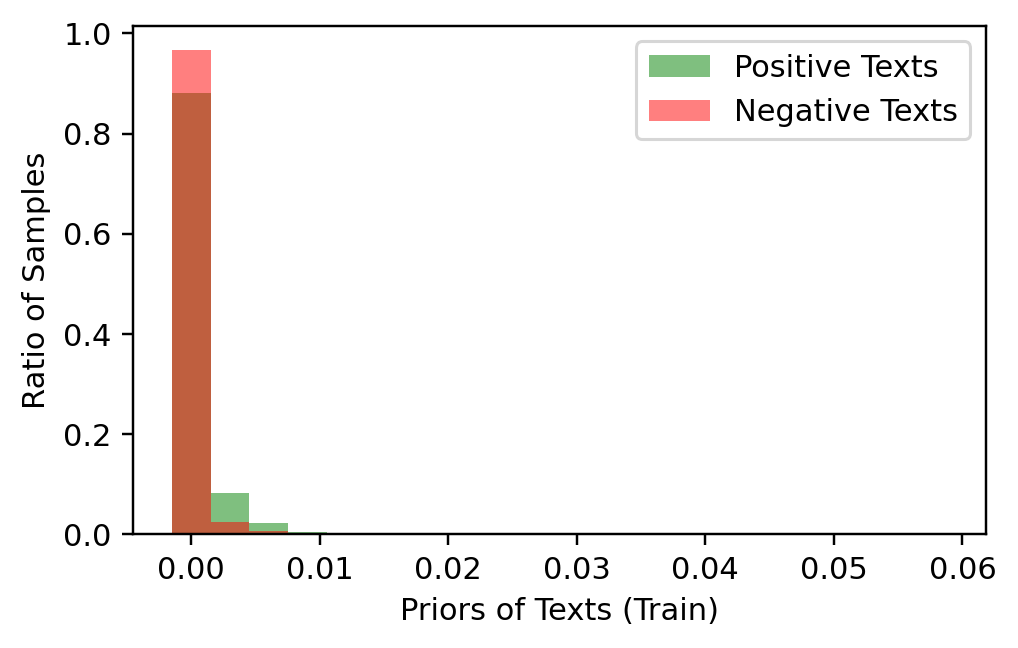} \\
\multicolumn{2}{c|}{SugarCrepe (Add)} & \multicolumn{2}{c}{VL-CheckList (Relation)} \\
\midrule
\includegraphics[width=0.19\textwidth]{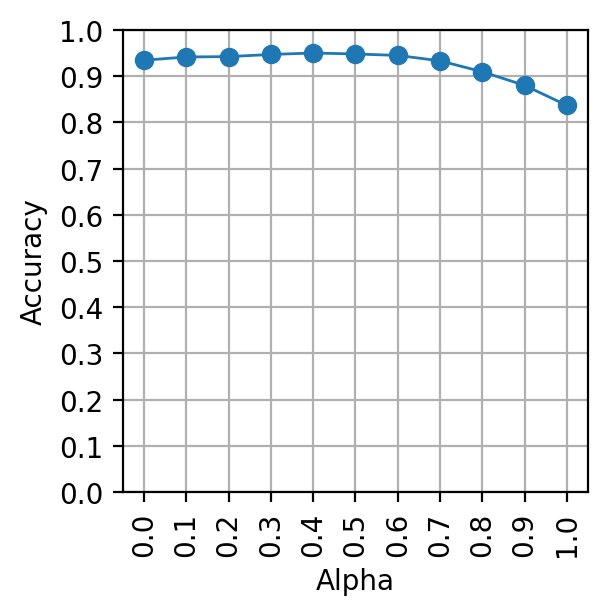} & \includegraphics[width=0.31\textwidth]{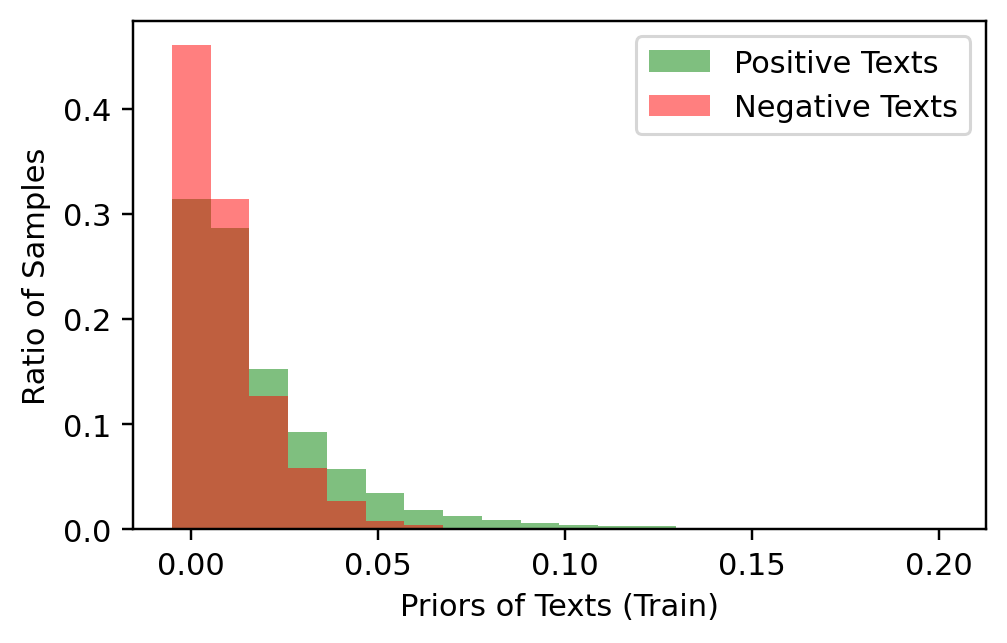} & \includegraphics[width=0.19\textwidth]{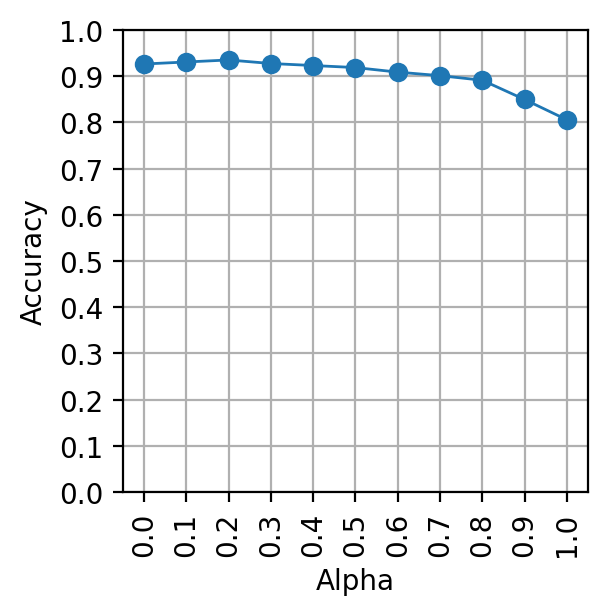} & \includegraphics[width=0.31\textwidth]{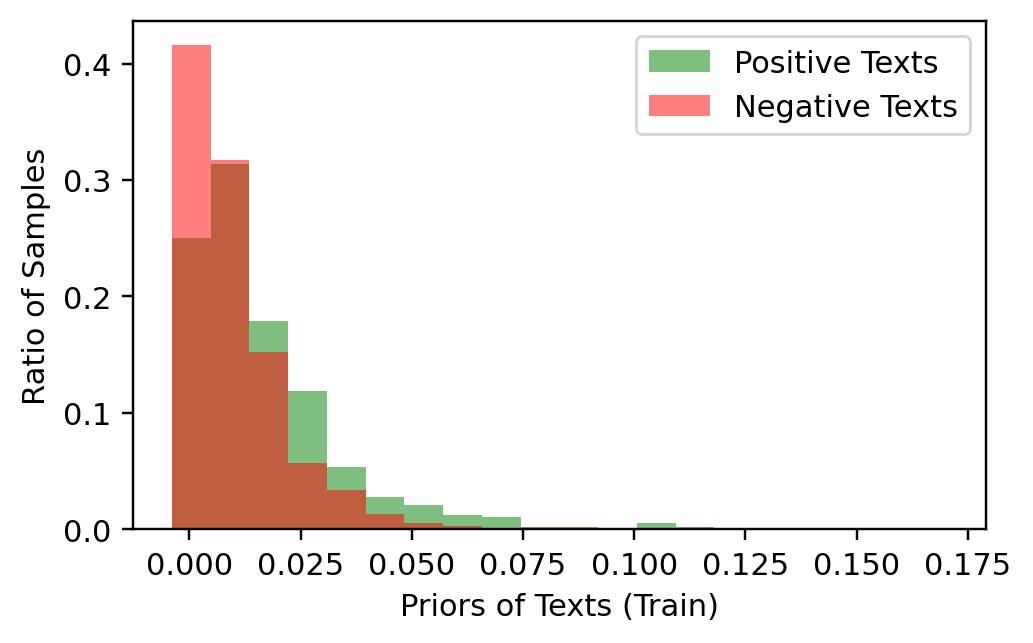} \\
\multicolumn{2}{c|}{SugarCrepe (Replace)} & \multicolumn{2}{c}{SugarCrepe (Swap)} \\
\bottomrule
\end{tabular}
}
\label{tab:datasets_plots}
\end{table*}

\clearpage

\section{Benchmark Visualization}
\label{app:dataset_visualization}
We include random samples from each benchmark in \autoref{tab:i2t_datasets}.

\begin{table}[h!]
\centering
\caption{\small {\bf Visualization of benchmarks.} ARO (VG-Relation/VG-Attribution/COCO-Order/Flickr30K-Order), Crepe (AtomFoils/Negate/Swap), VL-CheckList (Object/Attribute/Relation), SugarCrepe (Replace/Swap/Add) are constructed by generating hard negative captions for an image-text pair. On the other hand, each sample of Winoground and EqBen has two image-text pairs. }
\scalebox{0.62}{
\begin{tabular}{lcll}
\toprule
\midrule
Dataset & Image & \multicolumn{1}{l}{Positive Caption} & \multicolumn{1}{l}{Negative Caption(s)} \\ \midrule
 VG-Relation & \raisebox{-0.5\height}{\includegraphics[height=1.2cm]{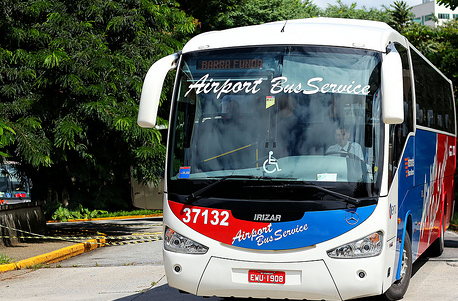}} & {\textcolor{black}{the bus is to the right of the trees}} & { \textcolor{purple}{the trees is to the right of the bus}}  \\ \addlinespace \midrule
 VG-Attribution & \raisebox{-0.5\height}{\includegraphics[height=1.2cm]{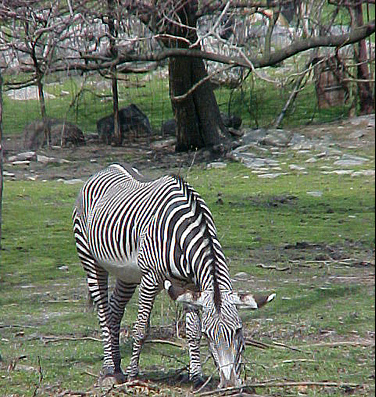}} & {\textcolor{black}{the striped zebra and the large tree}} & { \textcolor{purple}{the large zebra and the striped tree}}  \\ \addlinespace \midrule
 COCO-Order & \raisebox{-0.5\height}{\includegraphics[height=1.5cm]{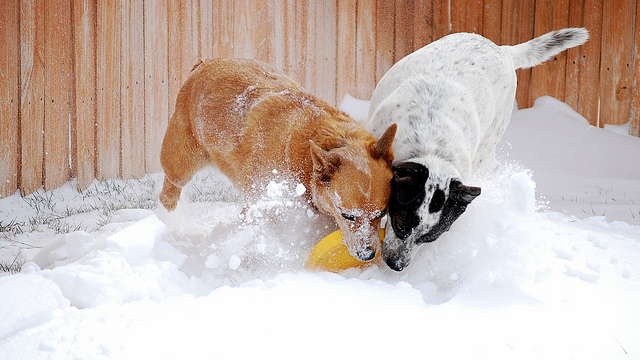}} & {\small \textcolor{black}{two dogs sharing a frisby in their mouth in the snow}} & \parbox{0.4\textwidth}{\scriptsize \textcolor{purple}{two frisby sharing a mouth in their snow in the dogs} \par \textcolor{purple}{in dogs the in frisby sharing two mouth their a snow} \par \textcolor{purple}{two dogs sharing in a frisby their mouth in snow the} \par \textcolor{purple}{a frisby in the snow two dogs sharing their mouth in}}  \\ \addlinespace \midrule
 Flickr30K-Order & \raisebox{-0.5\height}{\includegraphics[height=1.5cm]{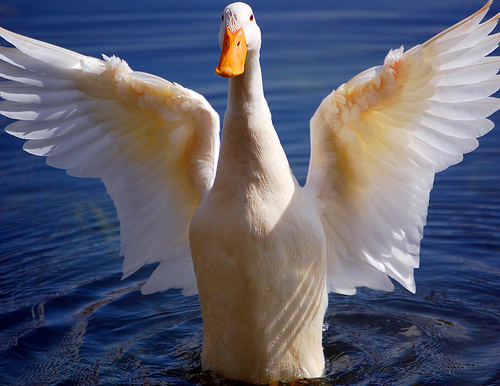}} & {\small \textcolor{black}{a white duck spreads its wings while in the water}} & \parbox{0.4\textwidth}{\scriptsize \textcolor{purple}{a white wings spreads its water while in the duck} \par \textcolor{purple}{a white duck the its wings while in water spreads} \par \textcolor{purple}{white a duck spreads its wings in while the water} \par \textcolor{purple}{while in the spreads its wings water a white duck}}  \\ \addlinespace \midrule
 \parbox{0.2\textwidth}{SugarCrepe \\ Add-Attribute} & \raisebox{-0.5\height}{\includegraphics[height=1.2cm]{images/datasets/sugarcrepe_add_att.jpg}} & {\textcolor{black}{They are going to serve pizza for lunch today.}} & {\textcolor{purple}{They are going to serve pizza topped with pineapple for lunch today.}}  \\ \addlinespace \midrule
  \parbox{0.2\textwidth}{SugarCrepe \\ Add-Object} & \raisebox{-0.5\height}{\includegraphics[height=1.2cm]{images/datasets/sugarcrepe_add_obj.jpg}} & {\textcolor{black}{A man kisses the top of a woman's head.}} & {\textcolor{purple}{A man kisses the top of a woman's head with a flower in his hand.}}  \\ \addlinespace \midrule
  \parbox{0.2\textwidth}{SugarCrepe \\ Replace-Attribute} & \raisebox{-0.5\height}{\includegraphics[height=1.2cm]{images/datasets/sugarcrepe_replace_att.jpg}} & {\textcolor{black}{A kid standing with a small suitcase on a street.}} & {\textcolor{purple}{A kid standing with a big suitcase on a street.}}  \\ \addlinespace \midrule
  \parbox{0.2\textwidth}{SugarCrepe \\ Replace-Object} & \raisebox{-0.5\height}{\includegraphics[height=1.2cm]{images/datasets/sugarcrepe.jpg}} & {\textcolor{black}{A duck floating in the water near a bunch of grass and rocks}} & {\textcolor{purple}{A swan floating in the water near a bunch of grass and rocks.}}  \\ \addlinespace \midrule
  \parbox{0.2\textwidth}{SugarCrepe \\ Replace-Relation} & \raisebox{-0.5\height}{\includegraphics[height=1.2cm]{images/datasets/sugarcrepe_replace_rel.jpg}} & {\textcolor{black}{A clock tower stands in front of a large mirrored sky scraper.}} & {\textcolor{purple}{A clock tower stands behind a large mirrored sky scraper.}}  \\ \addlinespace \midrule
  \parbox{0.2\textwidth}{SugarCrepe \\ Swap-Attribute} & \raisebox{-0.5\height}{\includegraphics[height=1.2cm]{images/datasets/sugarcrepe_swap_att.jpg}} & {\textcolor{black}{A tennis player is taking a swing on a red court.}} & {\textcolor{purple}{A red player is taking a swing on a tennis court.}}  \\ \addlinespace \midrule
  \parbox{0.2\textwidth}{SugarCrepe \\ Swap-Object} & \raisebox{-0.5\height}{\includegraphics[height=1.2cm]{images/datasets/sugarcrepe_swap_obj.jpg}} & {\textcolor{black}{A woman holding a game controller with a man looking on.}} & {\textcolor{purple}{A man holding a game controller with a woman looking on.}}  \\ \addlinespace \midrule
Crepe-AtomFoils & \raisebox{-0.5\height}{\includegraphics[height=1.5cm]{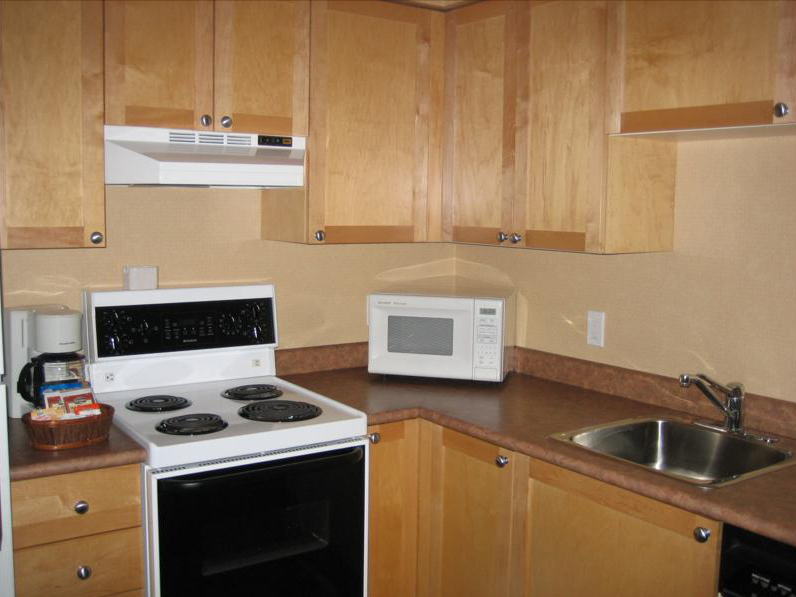}} & {\small \textcolor{black}{microwave in a kitchen, and sink in a kitchen.}} & \parbox{0.4\textwidth}{\scriptsize \textcolor{purple}{microwave in a cupboard, and sink in a kitchen} \par \textcolor{purple}{microwave in a bar, and sink in a kitchen} \par \textcolor{purple}{line in a kitchen, and sink in a kitchen} \par \textcolor{purple}{microwave in a kitchen, and shower in a kitchen} \par \textcolor{purple}{microwave in a kitchen, and tap in a kitchen}}  \\ \addlinespace \midrule
Crepe-Negate & \raisebox{-0.5\height}{\includegraphics[height=1.5cm]{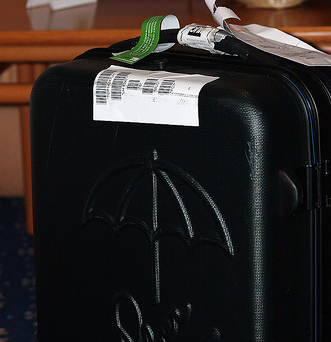}} & {\small \textcolor{black}{a chair next to a table, with the back of the chair visible.}} & \parbox{0.55\textwidth}{\scriptsize \textcolor{purple}{A chair is not next to a table, with the back of the chair visible} \par \textcolor{purple}{A chair next to a table, with the back not of the chair visible} \par \textcolor{purple}{A chair next to a table, with the back of the chair visible} \par \textcolor{purple}{A chair next to a table, with something of the chair visible. There is no back.} \par \textcolor{purple}{There is no chair next to a table, with the back of the chair visible}}  \\ \addlinespace \midrule
Crepe-Swap & \raisebox{-0.5\height}{\includegraphics[height=1.5cm]{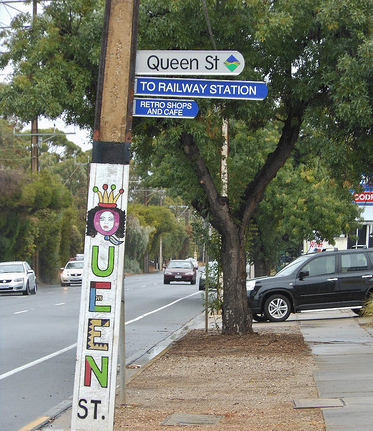}} & {\small \textcolor{black}{a car driving on a road with a line next to a tree.}} & \parbox{0.55\textwidth}{\scriptsize \textcolor{purple}{a car driving on a bright green leaves with a line next to a tree} \par \textcolor{purple}{a bright green leaves driving on a road with a line next to a tree} \par \textcolor{purple}{a car driving on a tree with a line next to a road} \par \textcolor{purple}{a car driving on a road with a line next to a white car} \par \textcolor{purple}{a car driving on a road with a line next to a street}}  \\ \addlinespace \midrule
\parbox{0.2\textwidth}{\small VL-CheckList \\ Relation (spatial)} & \raisebox{-0.5\height}{\includegraphics[height=1.2cm]{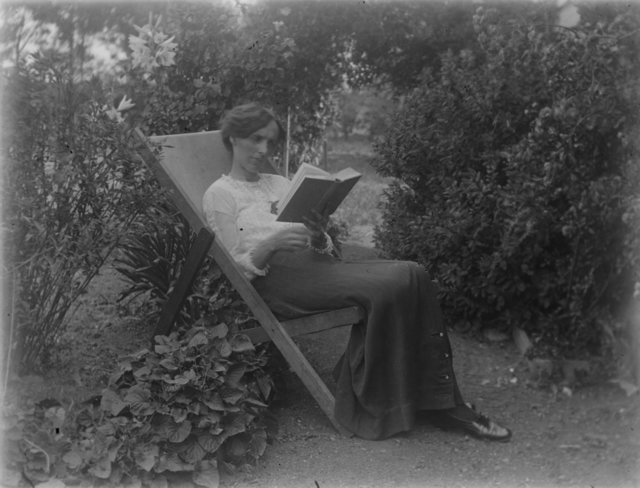}} & {\textcolor{black}{person read book}} & {\textcolor{purple}{person carry book}}  \\ \addlinespace \midrule
\parbox{0.2\textwidth}{\small VL-CheckList \\ Relation (action)} & \raisebox{-0.5\height}{\includegraphics[height=1.2cm]{images/datasets/vlchecklist_relation_spatial.png}} & {\textcolor{black}{sign near boy}} & {\textcolor{purple}{sign far from book}}  \\ \addlinespace \midrule
\midrule
\multirow{2}{*}{Winoground} & \raisebox{-0.5\height}{\includegraphics[height=1.2cm]{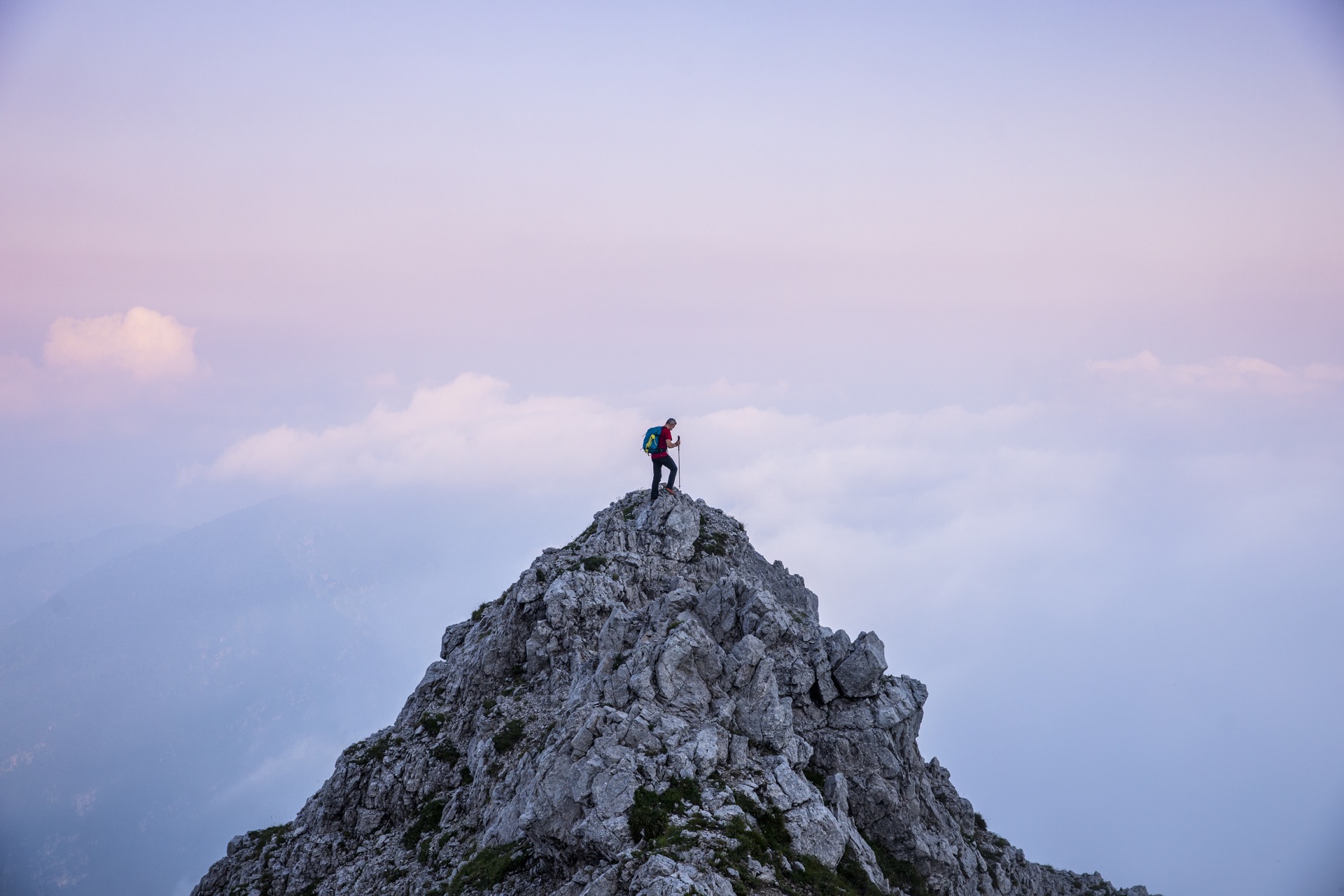}} & {\textcolor{black}{a  person on top of the world}} & {\textcolor{purple}{the world on top of a person}}  \\ \addlinespace
& \raisebox{-0.5\height}{\includegraphics[height=1.2cm]{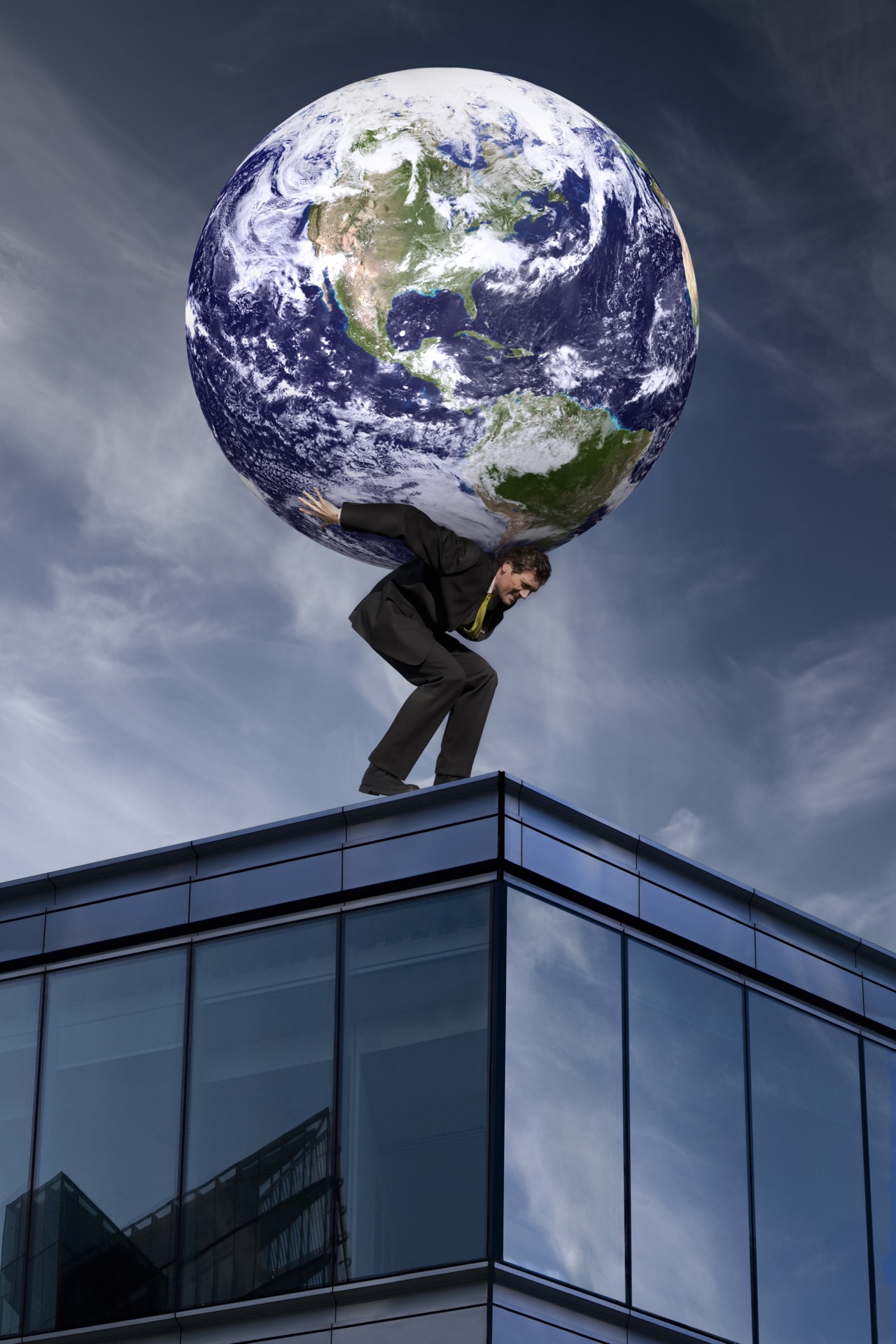}} & {\textcolor{black}{the world on top of a person}} & {\textcolor{purple}{a person on top of the world}}  \\ \addlinespace \midrule
\multirow{2}{*}{EqBen} & \raisebox{-0.5\height}{\includegraphics[height=1.2cm]{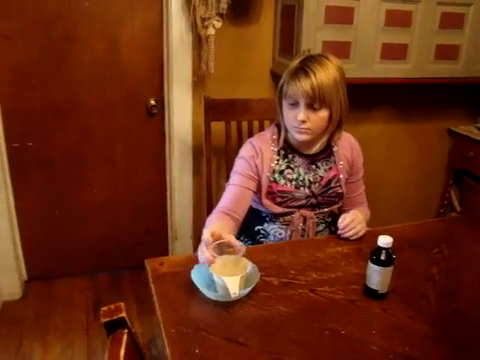}} & {\textcolor{black}{The person is touching the dish which is in front of him/her.}} & {\textcolor{purple}{The person is holding the dish which is in front of him/her.}}  \\ \addlinespace
& \raisebox{-0.5\height}{\includegraphics[height=1.2cm]{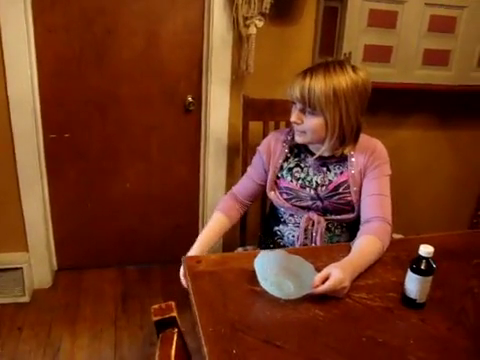}} & {\textcolor{black}{The person is holding the dish which is in front of him/her.}} & {\textcolor{purple}{The person is touching the dish which is in front of him/her.}}  \\ \addlinespace \midrule
\bottomrule
\end{tabular}
}
\label{tab:i2t_datasets}
\end{table}

\end{document}